\documentclass[]{fairmeta} 

\usepackage{amsmath,amsfonts,bm}

\def\eqref#1{equation~\ref{#1}}

\def\1{\bm{1}}

\DeclareMathAlphabet{\mathsfit}{\encodingdefault}{\sfdefault}{m}{sl}
\SetMathAlphabet{\mathsfit}{bold}{\encodingdefault}{\sfdefault}{bx}{n}

\usepackage[utf8]{inputenc}  
\usepackage[T1]{fontenc}     
\usepackage{hyperref}        
\usepackage{url}             
\usepackage{booktabs}        
\usepackage{amsfonts}        
\usepackage{nicefrac}        
\usepackage{microtype}       
\usepackage{graphicx}
\usepackage{multicol}
\usepackage{multirow}
\usepackage{amssymb}
\usepackage{xspace}
\usepackage{wrapfig}
\usepackage{enumitem}
\usepackage{colortbl}
\usepackage{caption}
\usepackage{lmodern} 
\usepackage[ruled,vlined]{algorithm2e}
\usepackage{subcaption}
\usepackage{xspace}
\usepackage{multicol}
\usepackage{multirow}
\usepackage{pifont}
\usepackage{tabularx}
\newcolumntype{R}{>{\raggedleft\arraybackslash}X}
\definecolor{lightblue}{RGB}{220,235,250}
\definecolor{lightgray}{gray}{0.91}

\usepackage[most,skins,theorems]{tcolorbox}
\usepackage{listings}
\newtcolorbox{promptbox}[1][]{
  enhanced, breakable,
  colback=gray!1,      
  colframe=gray!60,    
  coltitle=black,      
  boxrule=2pt,
  arc=10pt,
  left=6pt, right=6pt, top=6pt, bottom=6pt,
  title={#1}, fonttitle=\bfseries,
  attach boxed title to top left={yshift*=-3mm},
  boxed title style={colback=gray!10}
}

\tcbset{
  aibox/.style={
    width=\linewidth,
    top=8pt,
    bottom=4pt,
    colback=inftythink-red!15,
    colframe=inftythink-red,
    colbacktitle=inftythink-red!90!black,
    enhanced,
    center,
    attach boxed title to top left={yshift=-0.1in,xshift=0.15in},
    boxed title style={boxrule=0pt,colframe=white,},
  }
}
\newtcolorbox{AIbox}[2][]{aibox,title=#2,#1}

\usepackage{amsthm}
\theoremstyle{plain}

\theoremstyle{definition}

\theoremstyle{remark}

\newtcolorbox{casebox}[1][]{
  enhanced,
  breakable,                
  skin=standard,            
  colframe=black!75,        
  colback=black!2,          
  colbacktitle=black!75,    
  coltitle=white,           
  fonttitle=\bfseries\small,
  fontupper=\small,         
  sharp corners,            
  boxrule=0.8pt,            
  toptitle=1.5mm, bottomtitle=1.5mm, 
  left=2mm, right=2mm,      
  top=2mm, bottom=2mm,      
  title={#1},               
  colframe=black!75,        
}

\definecolor{myred}{RGB}{255, 0, 0}
\definecolor{myblue}{RGB}{0, 0, 255}
\definecolor{mygray}{RGB}{100, 100, 100} 
\newcommand{\diff}[2]{\textcolor{red}{+#1 / +#2}}
\newcommand{\res}[2]{#1 / #2}

\newcommand{\User}[1]{\vspace{0.8em}\noindent\textbf{User:} #1\par}
\newcommand{\Assistant}[1]{\vspace{0.8em}\noindent\textbf{Assistant:} #1\par}

\title{
CoVerRL: Breaking the Consensus Trap in Label-Free Reasoning via Generator-Verifier Co-Evolution
}

\author[1,2]{Teng Pan}
\author[1]{Yuchen Yan}
\author[1,2]{Zixuan Wang}
\author[2]{Ruiqing Zhang}
\author[1]{Guiyang Hou}
\author[1]{Wenqi Zhang}
\author[1]{Weiming Lu}
\author[1]{Jun Xiao}
\author[1 \dagger]{Yongliang Shen}

\affiliation[1]{Zhejiang University}
\affiliation[2]{Baidu Inc.}

\contribution[\dagger]{Corresponding authors}

\abstract{
Label-free reinforcement learning enables large language models to improve reasoning capabilities without ground-truth supervision, typically by treating majority-voted answers as pseudo-labels. However, we identify a critical failure mode: as training maximizes self-consistency, output diversity collapses, causing the model to confidently reinforce systematic errors that evade detection. We term this the consensus trap. To escape it, we propose CoVerRL, a framework where a single model alternates between generator and verifier roles, with each capability bootstrapping the other. Majority voting provides noisy but informative supervision for training the verifier, while the improving verifier progressively filters self-consistent errors from pseudo-labels. This co-evolution creates a virtuous cycle that maintains high reward accuracy throughout training. Experiments across Qwen and Llama model families demonstrate that CoVerRL outperforms label-free baselines by 4.7-5.9\% on mathematical reasoning benchmarks. Moreover, self-verification accuracy improves from around 55\% to over 85\%, confirming that both capabilities genuinely co-evolve.
}

\date{\today}
\metadata[Project Page]{\url{ https://zju-real.github.io/CoVerRL}}
\metadata[Code]{\url{https://github.com/ZJU-REAL/CoVerRL}}
\correspondence{\email{\{pt6,syl\}@zju.edu.cn}}

\begin{document}

\maketitle

\section{Introduction}
\label{sec:introduction}

Large Language Models (LLMs)~\citep{deepseek-ai2025deepseekr1,openai2024openai} have demonstrated remarkable capabilities in complex reasoning tasks, especially with recent advances in Reinforcement Learning with Verifiable Rewards (RLVR) that further amplify these abilities by optimizing models against ground-truth signals. However, the standard RLVR paradigm is heavily based on large amounts of high-quality labeled data or ground-truth verifiers, which are often unavailable or expensive to obtain. Consequently, enabling LLMs to self-evolve in a \textit{label-free} setting has emerged as a critical frontier in current research.

The dominant paradigm for label-free training employs majority voting as a proxy for correctness~\citep{prasad2025selfconsistency,zuo2025ttrl,liu2025ettrla}. Given a problem, the model generates multiple reasoning trajectories, and the most frequent answer becomes the pseudo-label for reward computation. The underlying assumption is that consistent answers across diverse reasoning paths likely reflect correct solutions. Notably, test-time reinforcement learning (TTRL)~\citep{zuo2025ttrl} reveal that majority voting remains effective even when pseudo-labels are incorrect through a ``lucky hit'' mechanism: samples whose answers differ from an incorrect pseudo-label still receive correct negative rewards. This explains why reward accuracy can remain high despite low label accuracy, providing viable training signals even on challenging benchmarks.

However, we identify a critical instability in this paradigm. Since rewards are computed against the majority answer, the training objective inherently encourages the model to produce more consistent outputs. As training progresses, output diversity systematically decreases and predictions concentrate around the majority answer. When this answer is incorrect, a destructive feedback loop emerges: the model confidently reinforces its own errors, diversity collapses further, and reward accuracy degrades. Figure~\ref{fig:intro} provides empirical evidence: 
\begin{wrapfigure}[21]{r}{0.5\textwidth}
    \centering
    \includegraphics[width=1\linewidth]{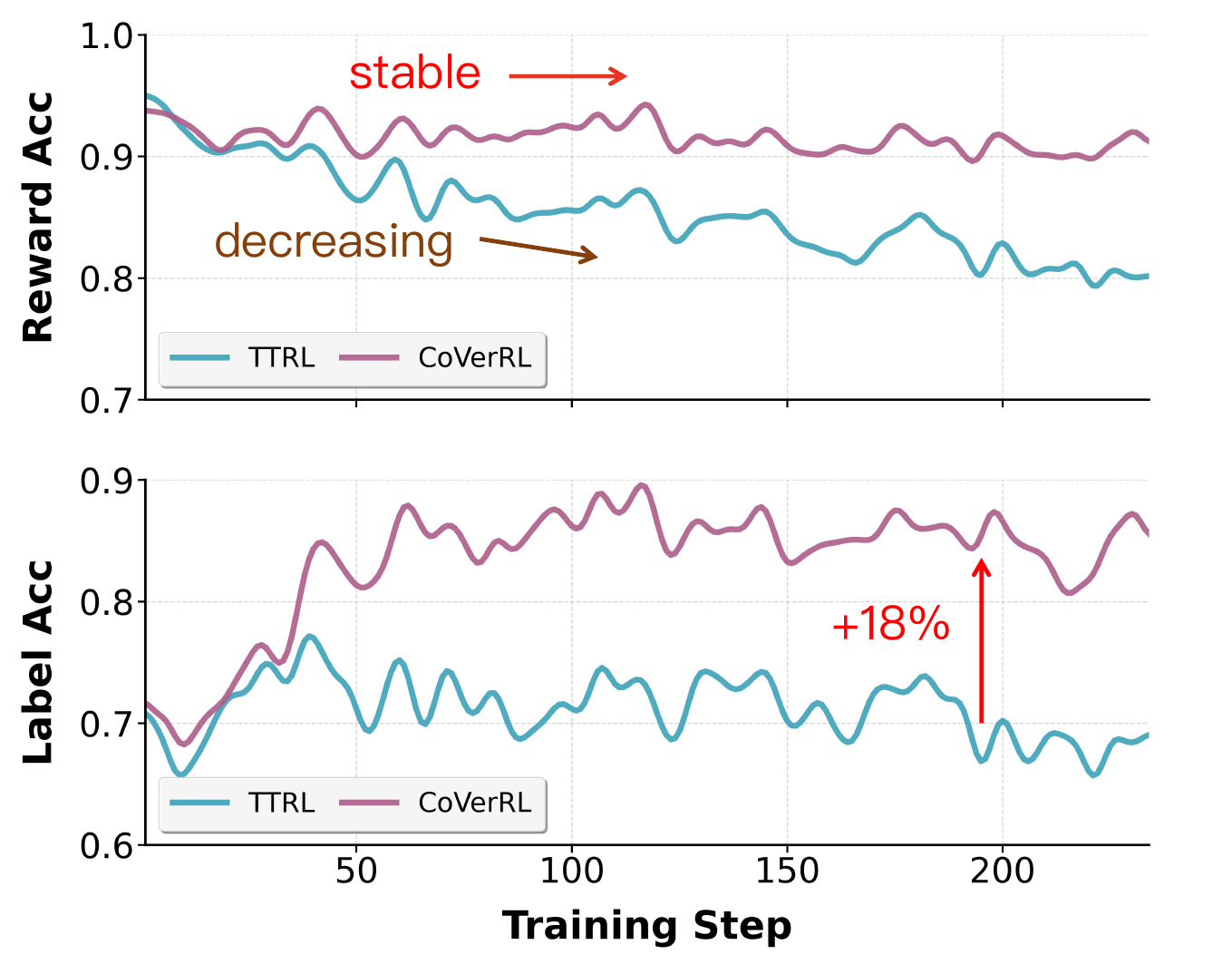}
    \caption{Training dynamics of TTRL and CoVerRL on Qwen3-1.7B-Base. CoVerRL maintains high reward accuracy while steadily improving label accuracy, whereas TTRL suffers from reward accuracy degradation as output diversity collapses.}
    \label{fig:intro}
\end{wrapfigure}
under TTRL, reward accuracy begins near 95\% but steadily drops to approximately 80\%, while label accuracy stagnates around 70\% with no improvement. Once the model converges to systematic errors with high confidence, pure consensus methods offer no mechanism for recovery. This reveals a fundamental limitation: majority voting captures consistency but cannot detect correctness. When models repeatedly produce identical wrong answers, a phenomenon known as self-consistent errors~\citep{tan2025tooa}, consensus-based rewards actively reinforce these mistakes.

A natural remedy is self-verification: if the model could judge correctness, self-consistent errors could be filtered before corrupting training. However, prior work has shown that LLMs without specific training cannot reliably detect their own reasoning errors~\citep{huang2024large, tyen2024llms}. Expecting a model to verify solutions it cannot correctly generate creates a circular dependency that severely limits self-correction without external supervision. The situation appears intractable: consensus provides training signal but cannot escape systematic errors, while verification could detect errors but cannot be trained without ground truth.

We resolve this tension through a key observation: majority voting and self-verification, while individually insufficient, can bootstrap each other's improvement. We propose \textbf{CoVerRL} (\textbf{Co}-evolving \textbf{Ver}ifier and generator via \textbf{R}einforcement \textbf{L}earning), a framework where a single model alternates between generator and verifier roles within a unified multi-turn RL process. The generator produces candidate solutions and establishes pseudo-labels via majority voting; the verifier scrutinizes reasoning processes to identify errors independent of answer frequency. The critical insight is bidirectional supervision: consensus-derived pseudo-labels provide contrastive signal for verifier training, while the improving verifier filters self-consistent errors from pseudo-labels. This creates a virtuous cycle where better verification yields cleaner labels, producing more accurate rewards that further improve both generation and verification. As shown in Figure~\ref{fig:intro}, CoVerRL maintains reward accuracy above 90\% throughout training, while label accuracy steadily increases from 75\% to 88\%, demonstrating the verifier's growing ability to escape the failure mode that plagues pure consensus methods.

We evaluate CoVerRL across three models on four reasoning benchmarks. CoVerRL consistently outperforms TTRL, achieving average improvements of 5.7\%, 5.9\%, and 4.7\% respectively across model scales. Beyond generation accuracy, CoVerRL produces capable verifiers as a byproduct: self-verification accuracy improves dramatically, from 55.8\% to 81.2\% on Qwen3-1.7B and from 54.0\% to 86.5\% on Qwen2.5-7B. This confirms that the two capabilities genuinely co-evolve rather than one merely supporting the other.

We summarize our contributions as follows:
\begin{itemize}[label={$\bullet$}, leftmargin=*] 
    \item We identify the consensus trap in majority voting based label-free RL, where diversity collapse causes reward accuracy degradation as models become overconfident in systematic errors, explaining why such methods eventually stagnate.
    \item We propose CoVerRL, a co-evolution framework that unifies generation and verification into a multi-turn RL process, enabling mutual bootstrapping where each capability supervises improvement of the other without external labels.
    \item We validate CoVerRL across Qwen and Llama model families, demonstrating 4-6\% improvements over label-free baselines on mathematical reasoning benchmarks while producing verifiers that generalize well to held-out evaluation.
\end{itemize}

\section{Related Work}
\label{sec:related_work}

\subsection{LLM Self-Evolution} To achieve label-free evolution, existing research has exploited the intrinsic confidence of LLMs as a supervision signal~\citep{prabhudesai2025maximizing,zhao2025learning,anonymous2025labelfree,li2025confidence}, leveraged self-consistency through majority voting to establish pseudo-labels~\citep{prasad2025selfconsistency,zuo2025ttrl,liu2025ettrla,zhou2025evolving,yu2025restrain,du2025testtimereinforcementlearninggui}, or employed self-verification mechanisms~\citep{yuan2025selfrewarding}. Another paradigm is self-play, where a single model simulates multiple roles and generates its own training data~\citep{zhao2025absolute,huang2025rzero,xia2025agent0,yang2025spell,guo2025genenv,wei2025training}. In this work, we leverage both majority voting and self-verification to co-evolve the generator and verifier, transcending the limitations of consensus trap and unreliable standalone verification.

\subsection{LLM Self-Verification and Self-Correction}
Recent advances have demonstrated that LLMs are capable of self-refinement when provided with external feedback~\citep{chen2023teachinga,liu2025agentbench,jain2024livecodebench,liu2025simple}. Concurrently, Several studies train standalone verifier models to bridge the verification gap~\citep{cobbe2021training,lightman2023lets,zhang2025generative,zhang2025lessons}. Beyond relying on external signals, many studies focus on intrinsic self-correction~\citep{miao2023selfcheck,lee2025revisea, ma2025s$^2$r, xiong2025selfrewardinga,yan2025s^3cmath}, which predominantly relies on either prompt engineering or two-stage paradigms combining Supervised Fine-Tuning (SFT) and Reinforcement Learning (RL). With the rapid progress in multi-turn RL~\citep{feng2025retool,jin2025searchr1,wang2025ragena,feng2025groupingroup}, recent studies have investigated training self-correction capabilities directly through a single reinforcement learning stage~\citep{liu2025trust, jiang2025pag}. In our work, we focus on enhancing self-verification and self-correction capabilities in a label-free paradigm, which has been rarely explored in previous studies.

\section{Method}
\label{sec:method_new}

In this section, we present CoVerRL, a framework that enables label-free reinforcement learning through the co-evolution of generation and verification capabilities. The key principle of CoVerRL is that a single policy model $\pi_\theta$ acts as two roles: a generator $\pi_\theta^{\text{gen}}$ and a verifier $\pi_\theta^{\text{ver}}$. Through their interaction, the model produces high-quality pseudo-labels (Section~\ref{sec:pseudo_label}) and achieves coordinated dual-role co-evolution (Section~\ref{sec:co_evolution}).

\begin{figure}[t]
    \centering
    \includegraphics[width=\linewidth]{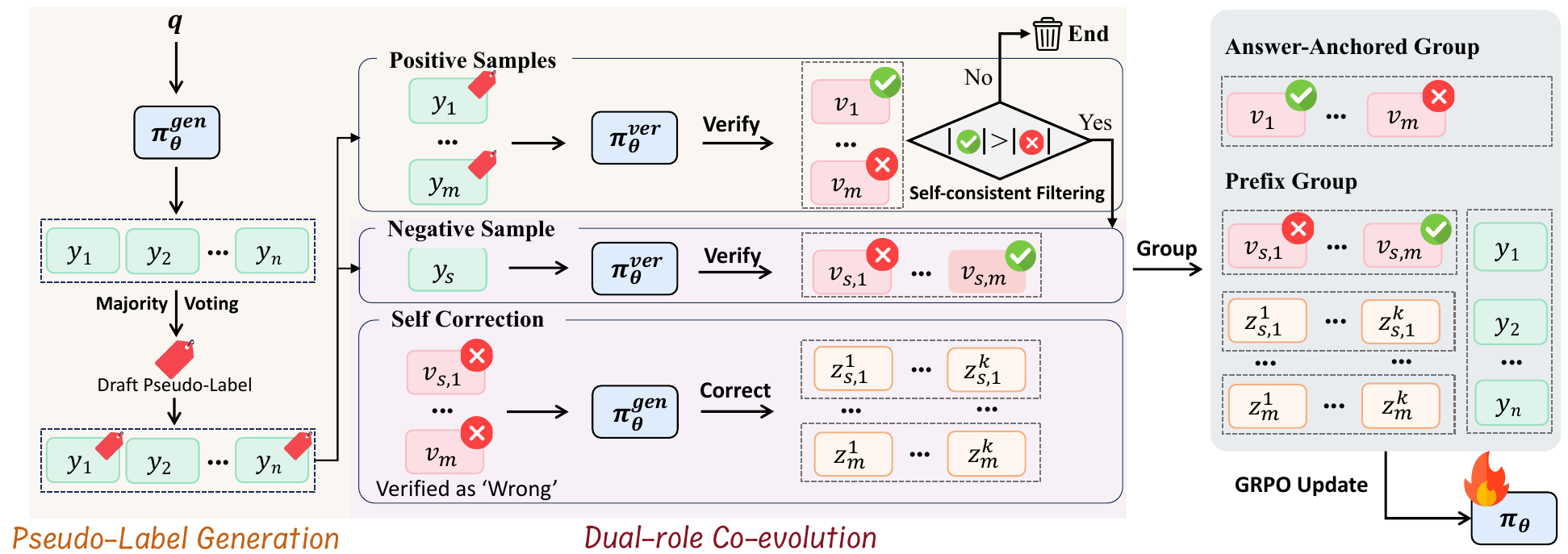}
    \caption{Overview of the CoVerRL framework. The pipeline illustrates the Pseudo-Label Generation process and the Online Dual-Role Co-Evolution strategy, facilitating a mutual bootstrapping process where generation and verification capabilities jointly evolve.}
    \label{fig:coverrl_framework}
\end{figure}

\subsection{Pseudo-Label Generation}
\label{sec:pseudo_label}
The key challenge in label-free RL is obtaining reliable supervision signals without ground truth. Majority voting suffers from diversity collapse, while self-verification lacks training signal. CoVerRL addresses this by combining both mechanisms: majority voting drafts initial pseudo-labels, and self-verification filters out self-consistent errors.

\paragraph{Majority Voting.}
Given a query $q$, the model acting as a generator samples $N$ reasoning paths $y_i \sim \pi_{\theta}^{\text{gen}}(\cdot \mid q)$, forming a collection of query-response pairs $Y = \{(q, y_i)\}_{i=1}^N$. Following~\citet{zuo2025ttrl}, we group these outputs and identify the most frequent answer as the draft pseudo-label $\hat{y}$. This partitions the outputs into a positive set $Y^+ = \{ (q, y_i) \in Y : \text{ans}(y_i) = \hat{y} \}$ and a negative set $Y^- = \{ (q, y_i) \in Y : \text{ans}(y_i) \neq \hat{y} \}$.

\paragraph{Self-consistent Verification Filtering.}
To filter self-consistent errors, the model switches to its verifier role $\pi_\theta^{\text{ver}}$ to scrutinize the candidate solutions. 
We define a maximum verification budget $M$ to maintain computational efficiency; if $|Y^+| > M$, we randomly sample $M$ tuples from $Y^+$ to form the verification set, otherwise include all tuples in $Y^+$. 
For each selected tuple $(q, y_j)$, the verifier produces a binary judgment $v_j \sim \pi_{\theta}^{\text{ver}}(\cdot \mid q, y_j)$, where $v_j \in \{0, 1\}$, and form the verification set $\mathcal{V}^+ = \{ (q, y_j, v_j) \}_{j=1}^{|\mathcal{V}^+|}$, where $|\mathcal{V}^+| = \min(|Y^{+}|, M)$. We then apply a consensus filter: the query $q$ and its associated sets are retained for training only if the majority of verifications in $\mathcal{V}^+$ are positive:
\begin{equation}
\small
    \sum_{j=1}^{|\mathcal{V}^+|} v_j > \frac{|\mathcal{V}^+|}{2}
\end{equation}
This filtering mechanism specifically targets the failure mode where the generator produces self-consistent errors; by leveraging $\pi_\theta^{\text{ver}}$ to identify flawed reasoning, we exclude such samples before they corrupt the training process.



\subsection{Online Dual-Role Co-Evolution}
\label{sec:co_evolution}

With filtered pseudo-labels established, we now describe how both roles are jointly optimized through multi-turn reinforcement learning.

\paragraph{Contrastive Verifier Training.}
To train the verifier without ground-truth labels, we construct contrastive examples from the derived pseudo-labels. The positive verification set $\mathcal{V}^+$ (targeting the majority answer $\hat{y}$) is obtained during the filtering stage. For negative examples, we select the answer $y_s$ with the lowest frequency from $Y^-$ and generate $|\mathcal{V}^+|$ verification paths to form the negative verification set $\mathcal{V}^- = \{ (q, y_s, v_j) \}_{j=1}^{|\mathcal{V}^+|} \sim \pi_{\theta}^{\text{ver}}(\cdot \mid q, y_s)$, where each $v_j$ targets the voted ``Incorrect'' status. Enforcing $|\mathcal{V}^-| = |\mathcal{V}^+|$ ensures balanced training, which we show in Appendix~\ref{app:proof} is equivalent to implicit preference optimization and induces beneficial curriculum effects.

\paragraph{Self-Correction.}
Beyond verification, we equip the model with error recovery capabilities by leveraging the interaction between its roles. For any triplet $(q, y, v) \in \mathcal{V}^+ \cup \mathcal{V}^-$ where the verifier identifies an error ($v=0$), the model switches back to its generator role $\pi_\theta^{\text{gen}}$. It is then triggered to produce $K$ revised solutions $\{z_k\}_{k=1}^K \sim \pi_{\theta}^{\text{gen}}(\cdot \mid q, y, v)$, conditioned on the query, the initial failed attempt, and the verifier's feedback. This structure allows the model to learn self-correction from its own internal critiques, extending the label-free paradigm to iterative refinement.

\paragraph{Reward Design.}
We define a reward function $r = r^f + r^a$ combining format and accuracy components. The format reward $r^f \in \{0, 1\}$ ensures outputs adhere to structural constraints. The accuracy reward is:

\begin{equation}
\small
    r^a = \mathbb{I}(o = \hat{o})
\end{equation}

where $o$ is the model's output and $\hat{o}$ is the corresponding pseudo-label. For generation, $\hat{o}$ is the consensus answer $\hat{y}$; for verification, $\hat{o}$ is the target judgment (``Correct'' for $\mathcal{V}^+$, ``Incorrect'' for $\mathcal{V}^-$).

\paragraph{Answer-Anchored GRPO.}
We optimize using Group Relative Policy Optimization (GRPO)~\citep{shao2024deepseekmath} with a modification for our multi-turn structure. Standard GRPO groups responses by query prefix. In our case, although verifications in $\mathcal{V}^+ = \{ (q, y_j, v_j) \}_{j=1}^{M}$ do not strictly originate from the same prefix, every $y_j$ targets the same answer $\hat{y}$. We thus propose \textbf{Answer-Anchored GRPO}: we group these $|\mathcal{V}^+|$ paths together as they share the same answer anchor $\hat{y}$; for generation and $\mathcal{V}^-$, we revert to standard prefix-based grouping. The advantage is computed as:
\begin{equation}
\small
    A_i = \frac{r_i - \text{mean}(\mathbf{r})}{\text{std}(\mathbf{r})}
\end{equation}

where $\mathbf{r}$ contains rewards within the appropriate group. This hybrid strategy isolates verification variance while maintaining stable generation optimization.

\paragraph{The Co-Evolution Cycle.}
The complete training loop proceeds as follows: (1) generate solutions and draft pseudo-labels via majority voting, (2) verify positive samples and filter self-consistent errors, (3) construct contrastive verification data, (4) collect self-correction trajectories, and (5) update the model with Answer-Anchored GRPO on all roles jointly. As training progresses, the verifier improves through contrastive learning, enabling better filtering of self-consistent errors. Cleaner pseudo-labels yield more accurate rewards, improving the generator. The improved generator produces higher-quality consensus labels, further benefiting verifier training. This bidirectional bootstrapping, visualized in Figure~\ref{fig:coverrl_framework}, enables both roles to co-evolve without external supervision.
The detailed algorithm flow is presented in Appendix~\ref{app:pseudo_code}.

\section{Experiments}
\label{sec:experiments}

\subsection{Environmental Setup}
\label{sec:environmental_setup}

\paragraph{Base Models.}
To verify the effectiveness of CoVerRL on diverse model architectures, we adopt three representative models from the Qwen and Llama families, including base and instruct models: Llama-3.2-3B-Instruct~\citep{grattafiori2024llama}, Qwen2.5-7B~\citep{qwen2025qwen25} and Qwen3-1.7B-Base~\citep{yang2025qwen3} with thinking mode.

\paragraph{Training and Evaluation Data.}
To test our method at scale, we use the large, standard MATH~\citep{hendrycks2021measuring} training set. For evaluation, we employ three in-domain mathematical benchmarks to assess reasoning capabilities: MATH500~\citep{hendrycks2021measuring}, AMC~\citep{li2024numinamath}, and AIME24~\citep{li2024numinamath}. Additionally, we include the GPQA-Diamond~\citep{rein2023gpqa} to evaluate the models' generalization on challenging out-of-domain question-answering tasks. 

\paragraph{Baseline and Implementation Details.} We mainly compare our method against TTRL~\citep{zuo2025ttrl}. For the TTRL baseline, we follow the practice in~\cite{zuo2025ttrl} by sampling 64 responses for majority voting and downsampling 32 responses for training. For our CoVerRL, at the first turn, we sample 32 responses for majority voting and downsample them to 16 for training; In later self-verification and self-correction, we utilize all samples for training to maximize data efficiency. About detailed description of the hyperparameters, please refer to Appendix~\ref{app:implementation_details}.

\paragraph{Evaluation Protocol.}
Following ~\citet{jiang2025pag}, we adopt a sequential inference strategy: the model first generates a solution, then verifies its correctness. If verification passes, inference terminates; otherwise, the model attempts a revision. We report both \textbf{Acc.@first} (first-attempt accuracy, equivalent to Pass@1) and \textbf{Acc.@final} (accuracy after potential self-correction), averaged over 16 different generations.
\subsection{Main Results}
\label{sec:main_results}

\begin{table*}[t]
\centering
\small
\renewcommand{\arraystretch}{1.2} 
\setlength{\tabcolsep}{8pt} 
\caption{Main results on mathematical reasoning and out-of-domain benchmarks. Each cell displays Acc.@first / Acc.@final. $\Delta$ indicates the improvement of CoVerRL over TTRL.}
\resizebox{\textwidth}{!}{%
\begin{tabular}{clccccc} 
\toprule
\textbf{Model} & \textbf{Method} & \textbf{MATH500} & \textbf{AMC} & \textbf{AIME24} & \textbf{GPQA} & \textbf{Average} \\
\midrule

\multirow{4}{*}{\textbf{Qwen3-1.7B-Base}} 
& Base Model & \res{53.5}{53.3} & \res{24.6}{24.5} & \res{3.8}{3.3} & \res{27.5}{27.3} & \res{27.4}{27.1} \\
& TTRL & \res{65.1}{65.0} & \res{31.1}{30.9} & \res{5.2}{5.2} & \res{30.9}{30.7} & \res{33.1}{33.0} \\
& CoVerRL (Ours) & \res{\textbf{69.0}}{\textbf{71.9}} & \res{\textbf{36.0}}{\textbf{38.6}} & \res{\textbf{9.8}}{\textbf{10.6}} & \res{\textbf{32.9}}{\textbf{33.6}} & \res{\textbf{36.9}}{\textbf{38.7}} \\
& \textbf{$\Delta$} & \diff{3.9}{6.9} & \diff{4.9}{7.7} & \diff{4.6}{5.4} & \diff{2.0}{2.9} & \diff{3.8}{5.7} \\
\midrule

\multirow{4}{*}{\textbf{Llama-3.2-3B-Instruct}} 
& Base Model & \res{42.7}{41.0} & \res{17.0}{15.7} & \res{4.6}{5.0} & \res{26.9}{26.1} & \res{22.8}{22.0} \\
& TTRL & \res{52.6}{52.2} & \res{23.8}{23.3} & \res{13.8}{14.0} & \res{29.8}{28.2} & \res{30.0}{29.4} \\
& CoVerRL (Ours) & \res{\textbf{55.9}}{\textbf{59.3}} & \res{\textbf{28.3}}{\textbf{32.2}} & \res{\textbf{16.3}}{\textbf{16.9}} & \res{\textbf{32.3}}{\textbf{32.6}} & \res{\textbf{33.2}}{\textbf{35.3}} \\
& \textbf{$\Delta$} & \diff{3.3}{7.1} & \diff{4.5}{8.9} & \diff{2.5}{2.9} & \diff{2.5}{4.4} & \diff{3.2}{5.9} \\
\midrule

\multirow{4}{*}{\textbf{Qwen2.5-7B}} 
& Base Model & \res{50.1}{51.4} & \res{25.5}{26.4} & \res{5.2}{6.5} & \res{29.9}{29.7} & \res{27.7}{28.5} \\
& TTRL & \res{73.8}{74.2} & \res{42.2}{42.2} & \res{12.7}{12.5} & \res{35.8}{35.6} & \res{41.1}{41.1} \\
& CoVerRL (Ours) & \res{\textbf{76.8}}{\textbf{79.6}} & \res{\textbf{47.6}}{\textbf{49.2}} & \res{\textbf{14.6}}{\textbf{17.1}} & \res{\textbf{36.2}}{\textbf{37.2}} & \res{\textbf{43.8}}{\textbf{45.8}} \\
& \textbf{$\Delta$} & \diff{3.0}{5.4} & \diff{5.4}{7.0} & \diff{1.9}{4.6} & \diff{0.4}{1.6} & \diff{2.7}{4.7} \\

\bottomrule
\end{tabular}%
}

\label{tab:main_generator_results}
\end{table*}

Table~\ref{tab:main_generator_results} presents the main results. CoVerRL consistently outperforms TTRL across all models and benchmarks, achieving average improvements of 5.7\%, 5.9\%, and 4.7\% in Acc.@final for the three models respectively.

\paragraph{Improved Generation Quality.}
CoVerRL improves Acc.@first over TTRL by 3.8\%, 3.2\%, and 2.7\% on average across the three models, demonstrating that the co-evolution mechanism enhances the generator itself, not merely the correction process. This confirms that filtering self-consistent errors produces cleaner training signals that benefit generation quality directly.

\paragraph{Effective Self-Correction.}
The gap between Acc.@first and Acc.@final widens substantially under CoVerRL compared to TTRL. On AMC, CoVerRL achieves Acc.@final improvements of 7.7\%, 8.9\%, and 7.0\% over TTRL, significantly larger than the Acc.@first gains. This indicates that the verifier trained through co-evolution can reliably identify errors and guide effective corrections, whereas TTRL's self-correction provides minimal benefit due to unreliable verification.

\begin{wraptable}[10]{r}{0.5\textwidth}  
\centering
\small
\renewcommand{\arraystretch}{1.2} 
\vspace{-1em}
\caption{Self-verification accuracy on four benchmarks.}
\resizebox{\linewidth}{!}{           
    \begin{tabular}{lcccc}
    \toprule
    \textbf{Model} & \textbf{MATH500} & \textbf{AMC} & \textbf{AIME24} & \textbf{GPQA} \\
    \midrule
    Qwen3-1.7B-Base & 55.8 & 28.5 & 11.7 & 38.7 \\
    + \textbf{CoVerRL} & \textbf{81.2} & \textbf{64.4} & \textbf{48.9} & \textbf{57.9} \\
    \midrule
    Llama-3.2-3B-Instruct & 57.0 & 50.9 & 55.8 & \textbf{56.9} \\
    + \textbf{CoVerRL} & \textbf{77.2} & \textbf{76.2} & \textbf{75.6} & 55.2 \\
    \midrule
    Qwen2.5-7B & 54.0 & 30.9 & 12.9 & 35.7 \\
    + \textbf{CoVerRL} & \textbf{86.5} & \textbf{70.9} & \textbf{58.8} & \textbf{58.4} \\
    \bottomrule
    \end{tabular}
}
\label{tab:self_verification}
\end{wraptable}

\paragraph{Out-of-Domain Generalization.}
On GPQA-Diamond, a challenging out-of-domain benchmark, CoVerRL maintains consistent improvements. Notably, while base models and TTRL often suffer degradation during revision (Acc.@final $<$ Acc.@first), CoVerRL reverses this trend across all models. This suggests that the verification capability learned through co-evolution transfers beyond the training distribution.


\paragraph{Verifier Quality.}
Table~\ref{tab:self_verification} reports self-verification accuracy. CoVerRL dramatically improves verification capability: Qwen3-1.7B-Base improves from 28.5\% to 64.4\% on AMC, and Qwen2.5-7B improves from 54.0\% to 86.5\% on MATH500. These gains confirm that both generation and verification genuinely co-evolve rather than one merely supporting the other.

\subsection{Analysis}
\label{sec:analysis}


\paragraph{Verifier and Generator Co-Evolve Mutually.}

\begin{figure*}[h]
    \centering
    \includegraphics[width=1\textwidth]{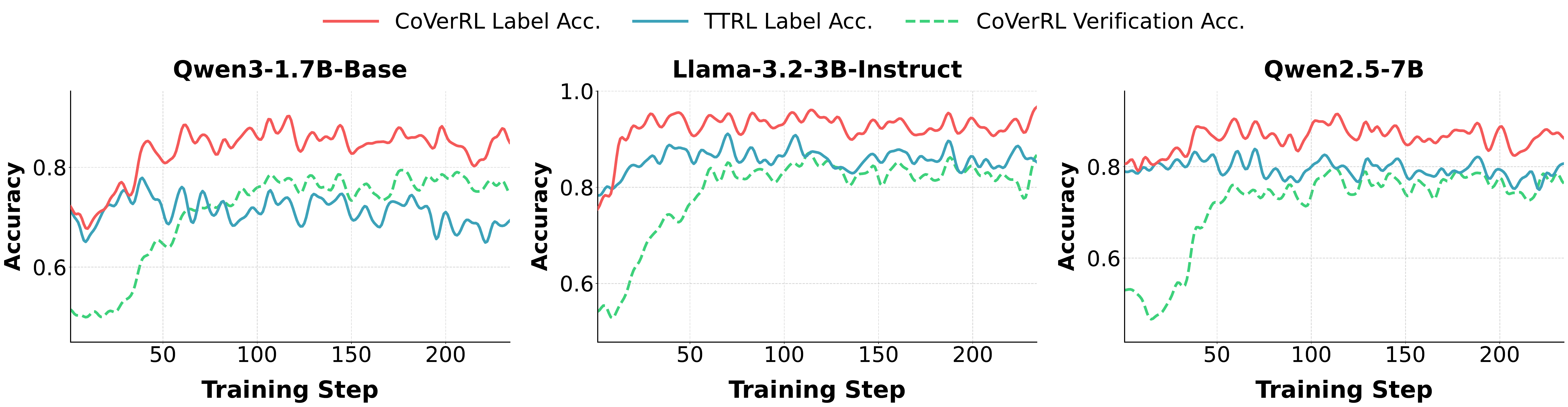}
    \caption{Co-evolution dynamics across three model backbones. As training progresses, verification accuracy improves through contrastive learning, which in turn enables better filtering of self-consistent errors and leads to higher label accuracy than TTRL.}
    \label{fig:how_work}
\end{figure*}

\begin{figure}[t]
  \centering
  \begin{minipage}[t]{0.49\linewidth}
    \centering
    \includegraphics[width=\linewidth]{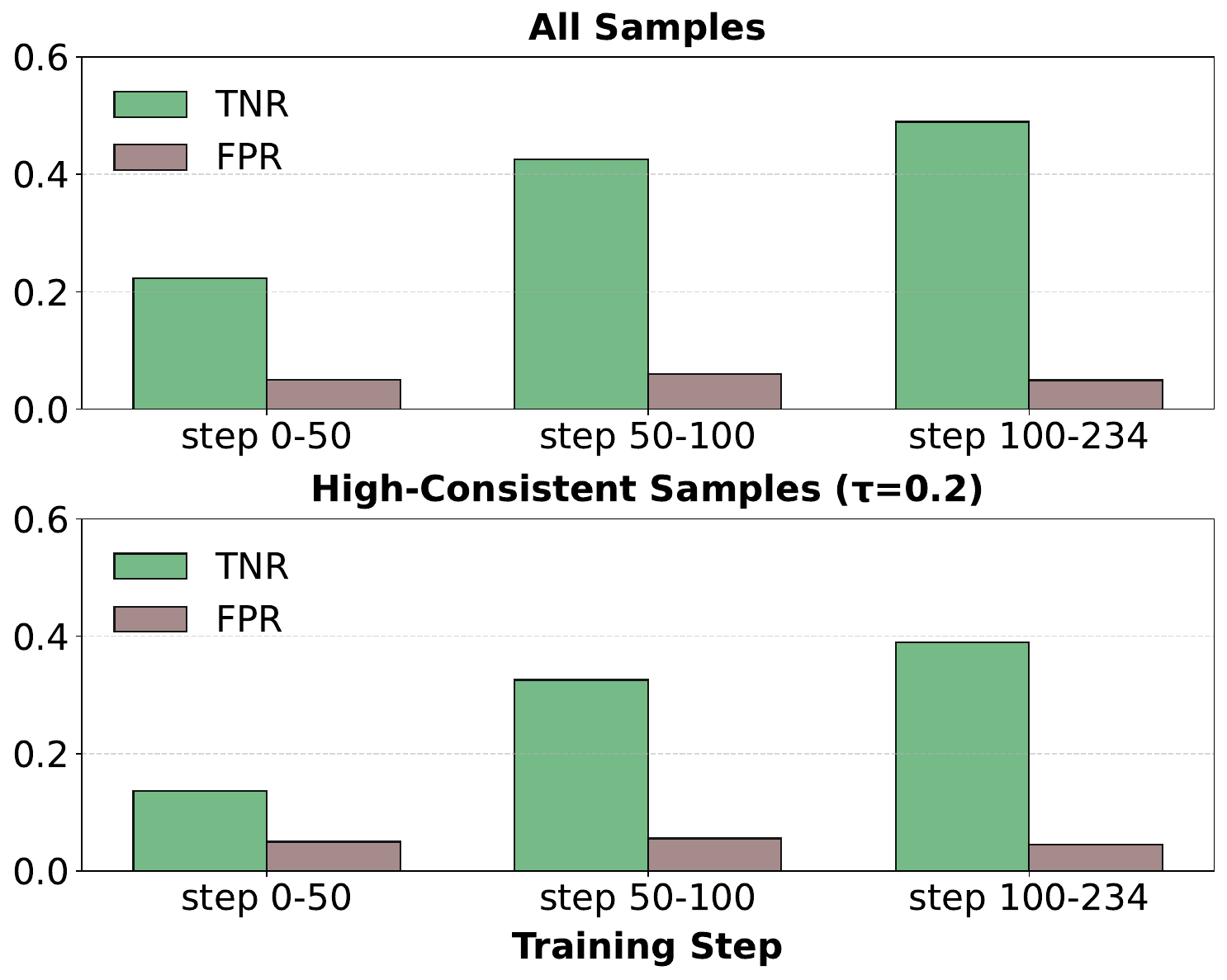}
    \captionof{figure}{Evolution of the verification capability of the Qwen3-1.7B-Base. We report the True Negative Rate (TNR) and the False Positive Rate (FPR).}
    \label{fig:filter_metric}
  \end{minipage}\hfill
  \begin{minipage}[t]{0.49\linewidth}
    \centering
    \includegraphics[width=\linewidth]{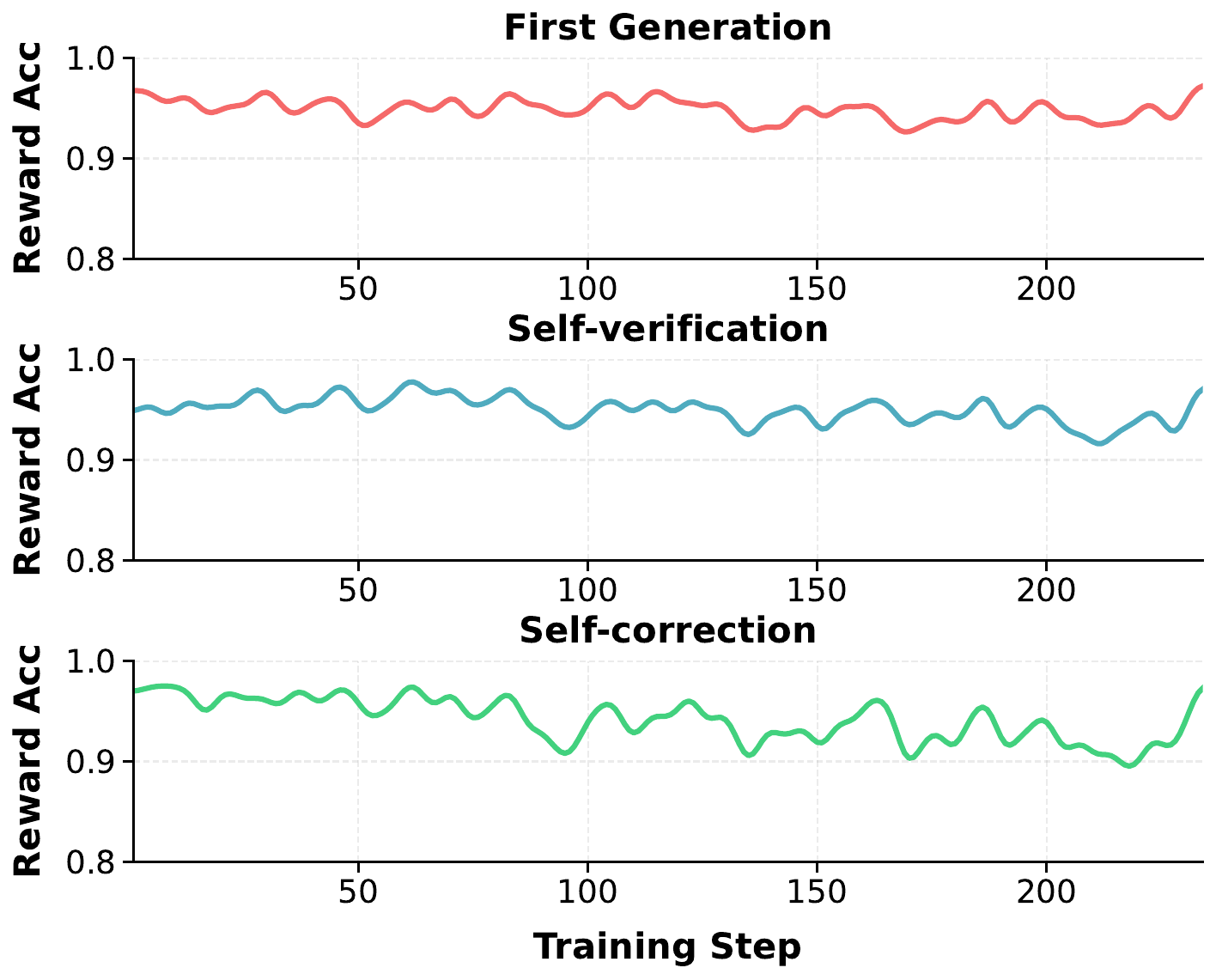}
    \captionof{figure}{Reward accuracy across generation, verification, and self-correction roles during training on Llama-3.2-3B-Instruct.}
    \label{fig:llama_reward_acc}
  \end{minipage}
\end{figure}

Figure~\ref{fig:how_work} reveals the co-evolution dynamics across three models. In early training, verifier accuracy (green dashed line) starts low due to the lack of verification-specific supervision, resulting in label accuracy comparable to TTRL. As training progresses, the verifier improves by learning from pseudo-labels that, while noisy, provide contrastive signal between majority and minority answers. This improved verification capability enables better filtering of self-consistent errors. Consequently, CoVerRL's label accuracy (red line) steadily increases and eventually surpasses TTRL (blue line) by a significant margin across all three model backbones. The consistent pattern across different architectures and scales confirms that both capabilities bootstrap each other through the bidirectional supervision mechanism rather than one merely supporting the other.
For a clearer demonstration of the verifier’s evolution, Figure~\ref{fig:filter_metric} reports the True Negative Rate (TNR), the ratio of incorrect samples correctly filtered, and the False Positive Rate (FPR), the proportion of correct samples mistakenly discarded. The left panel shows that TNR steadily climbs from approximately 22\% to nearly 50\%, while the FPR remains stable at around 5\%, indicating the verifier increasingly prunes erroneous paths while maintaining a minimal false-alarm rate for correct solutions.

\paragraph{Reward Accuracy Remains High Across All Roles.}
Figure~\ref{fig:llama_reward_acc} tracks reward accuracy for generation, verification, and self-correction throughout training on Llama-3.2-3B-Instruct. All three roles maintain reward accuracy above 90\%, indicating that the pseudo-labels derived from the co-evolution process provide reliable supervision signals for each role. 
The sustained high reward accuracy across all roles demonstrates that the bidirectional bootstrapping mechanism successfully prevents the degradation identified in Section~\ref{sec:introduction}.
yielding both a marked improvement in Acc.@first and a consistently higher Acc.@final.

\begin{figure}[t]
    \centering
    \includegraphics[width=0.9\linewidth]{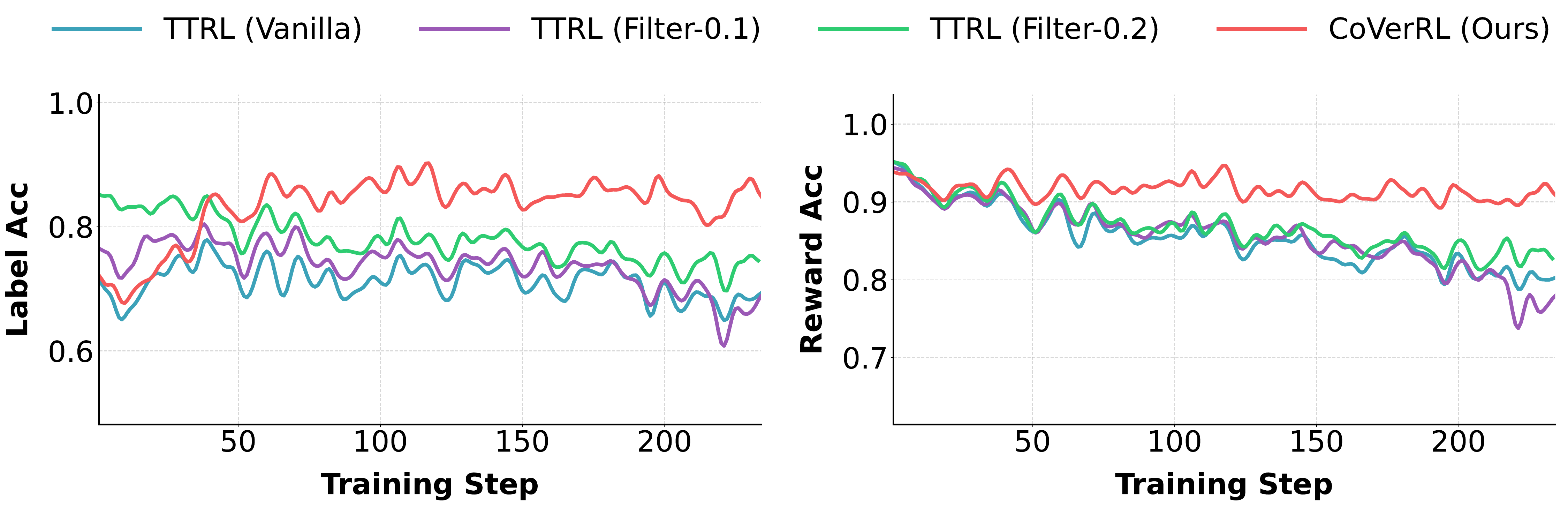}
\caption{Comparison of CoVerRL against naive consistency filtering on Qwen3-1.7B-Base. Naive filtering ($\tau \in \{0.1, 0.2\}$) improves label accuracy initially but converges to TTRL baseline, while failing to improve reward accuracy. CoVerRL achieves sustained gains on both metrics.}
    \label{fig:high_errors}
\end{figure}

\paragraph{Verification Filtering Outperforms Naive Consistency Filtering.}
Since samples with lower majority ratios are more likely to be incorrect, a natural alternative to verification-based filtering is to simply discard samples below a consistency threshold. We compare CoVerRL against TTRL with naive filtering thresholds $\tau \in \{0.1, 0.2\}$ in Figure~\ref{fig:high_errors}. The left panel shows label accuracy dynamics: while naive filtering initially improves label accuracy by removing low-confidence predictions, its advantage diminishes as training progresses and model outputs become increasingly consistent due to the RL objective, eventually converging to vanilla TTRL performance. In contrast, CoVerRL's label accuracy continues to improve throughout training, reaching approximately 90\% compared to TTRL's 70\%. The right panel reveals a more fundamental distinction in reward accuracy: naive filtering provides no improvement over vanilla TTRL because it cannot detect errors that happen to achieve high consensus. CoVerRL maintains reward accuracy above 90\% by specifically targeting self-consistent errors through reasoning-based verification. 
As shown in the right panel of Figure~\ref{fig:filter_metric}, for samples with majority ratio larger than 0.2, the TNR reaches around 40\% after 100 steps, confirming the verifier's strong capability to prune confident errors missed by naive filtering.


\begin{wraptable}[15]{r}{0.5\textwidth}
\centering
\small
\setlength{\tabcolsep}{8pt}  
\vspace{-1em}
\caption{Performance on RewardBench. Scores marked with $^*$ are from the RewardBench leaderboard.}
\begin{tabular}{lc}
\toprule
\textbf{Model} & \textbf{Score} \\
\midrule
Qwen3-1.7B-Base & 46.8\phantom{$^*$} \\
Llama-3.2-3B-Instruct & 58.6\phantom{$^*$} \\
Qwen2.5-7B & 49.7\phantom{$^*$} \\
\midrule
Meta-Llama-3-70B-Instruct & 66.2$^*$ \\
gpt-4-turbo-2024-04-09 & 67.3$^*$ \\
gpt-4o-2024-05-13 & 72.5$^*$ \\
\midrule
\textbf{Qwen3-1.7B-Base + CoVerRL} & 56.4\phantom{$^*$} \\
\textbf{Llama-3.2-3B-Instruct + CoVerRL} & 70.0\phantom{$^*$} \\
\textbf{Qwen2.5-7B + CoVerRL} & 60.9\phantom{$^*$} \\
\bottomrule
\end{tabular}
\label{tab:rewardbench}
\end{wraptable}

\paragraph{Verification Capability Generalizes Beyond Training Distribution.}
To assess whether the learned verification capability transfers beyond the model's own generated responses, we evaluate the trained verifiers on RewardBench~\citep{lambert2024rewardbencha} mathprm. As shown in Table~\ref{tab:rewardbench}, CoVerRL-enhanced models substantially outperform their base versions across all three architectures. Most notably, Llama-3.2-3B-Instruct improves from 58.6 to 70.0, surpassing the significantly larger Meta-Llama-3-70B-Instruct (66.2) and GPT-4-Turbo (67.3). This strong performance suggests that CoVerRL trains genuine reasoning-based verification capabilities and that the co-evolution process induces transferable skills for evaluating solution quality.

\begin{figure}[h]
    \centering
    \includegraphics[width=0.9\linewidth]{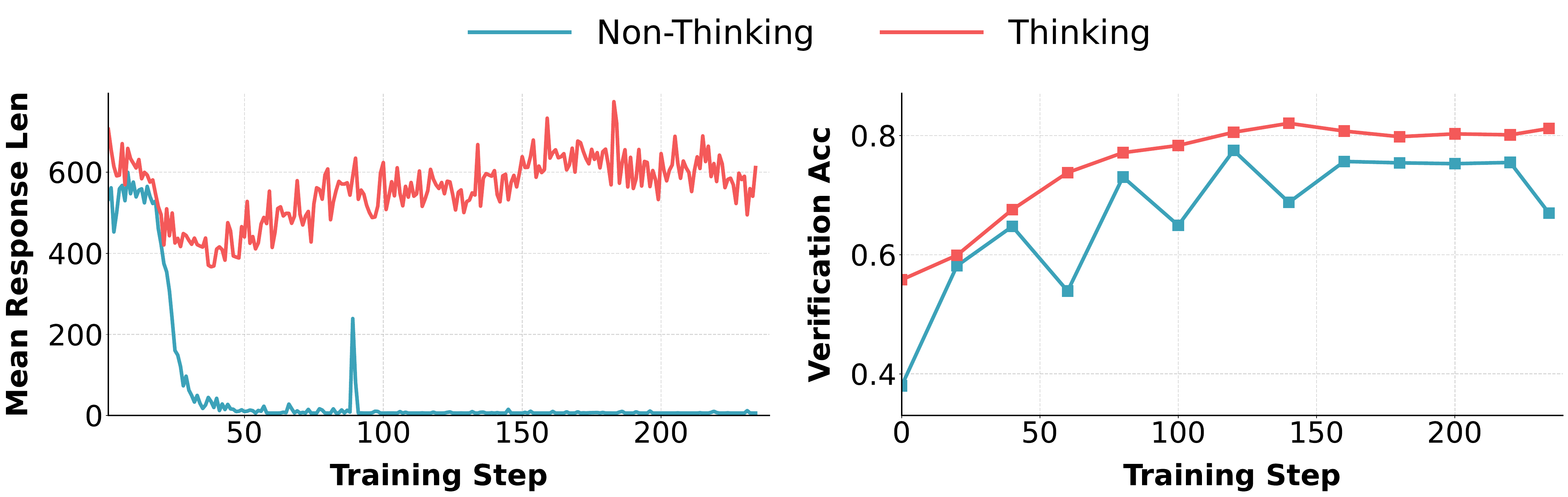}
    \caption{Comparison of training response length and validation accuracy on MATH500 between the \textit{Thinking mode} and \textit{Non-thinking mode} on Qwen3-1.7B-Base. Non-thinking mode suffers from length collapse (left), leading to inferior verification accuracy (right).}
    \label{fig:thinking_vs_nonthinking}
\end{figure}

\paragraph{Reasoning Capability is Essential for Effective Verification.}
To validate that verification quality depends on substantive reasoning rather than superficial pattern matching, we compare thinking mode versus non-thinking mode on Qwen3-1.7B-Base in Figure~\ref{fig:thinking_vs_nonthinking}. The non-thinking mode removes the extended reasoning process during verification. Without this reasoning capacity, the model rapidly collapses to trivial outputs such as ``The answer is correct'' or ``The answer is wrong,'' with mean response length dropping from over 500 tokens to near zero within 100 training steps. This length collapse severely limits the verifier's discriminative ability, resulting in significantly lower verification accuracy compared to thinking mode throughout training. The result confirms that CoVerRL's effectiveness stems from the verifier's ability to scrutinize and reason about solution processes, not merely from adding a binary verification step.
\subsection{Ablation Study}
\label{sec:ablation_study}

\begin{figure}[h]
    \centering
    \includegraphics[width=0.9\linewidth]{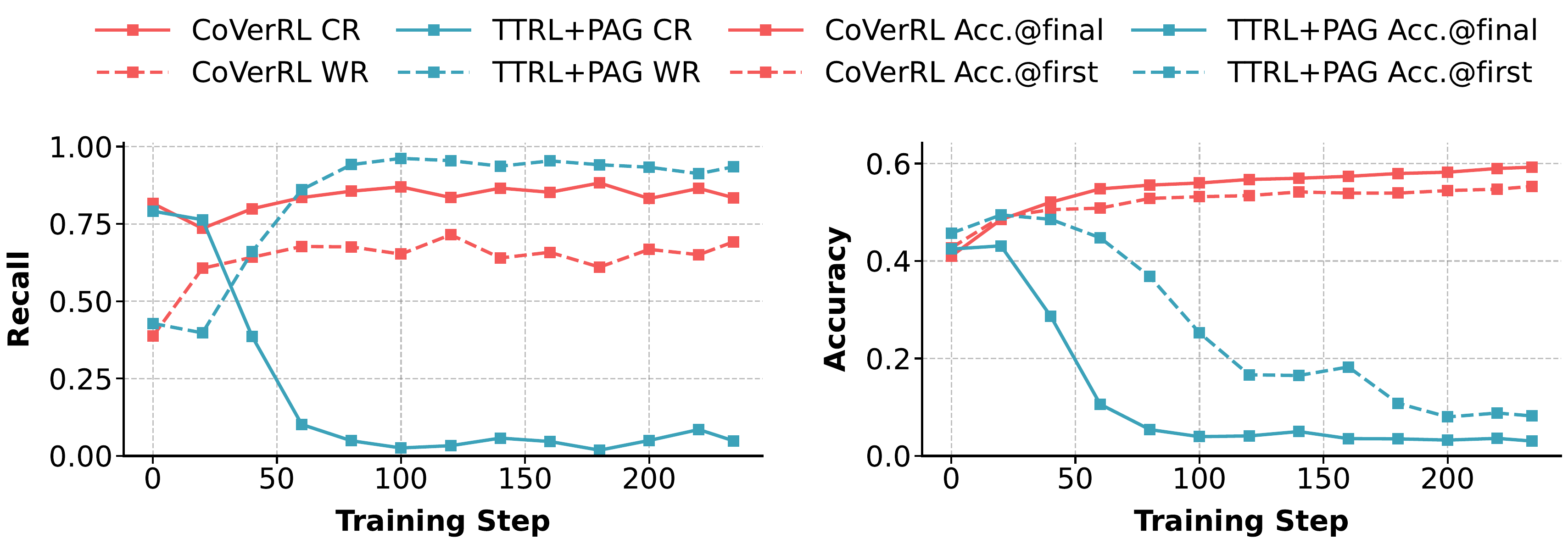}
    \caption{Comparison of TTRL + PAG and CoVerRL on Llama-3.2-3B-Instruct. TTRL + PAG exhibits verifier collapse with Correct Recall dropping to near zero (left), causing accuracy degradation (right). CoVerRL maintains stable training through balanced sampling.}
    \label{fig:analysis_balance}
\end{figure}

\paragraph{Balanced Training Prevents Verifier Collapse.}
We investigate whether existing multi-turn RL methods can be directly combined with majority voting for label-free training. We construct a baseline named TTRL + PAG, which applies PAG's verify-then-revise workflow~\citep{jiang2025pag} using majority-voted pseudo-labels instead of ground truth, while keeping other components in PAG (PPO optimization, turn-independent advantage normalization) unchanged. Figure~\ref{fig:analysis_balance} compares this baseline against CoVerRL on Llama-3.2-3B-Instruct. TTRL + PAG suffers catastrophic collapse: Correct Recall (the proportion of correct answers identified as correct) drops to near zero while Wrong Recall (the proportion of incorrect answers identified as wrong) surges to nearly 100\%. This indicates the verifier degenerates into rejecting all solutions indiscriminately. The failure stems from information asymmetry in the training data: positive samples that reach the majority answer tend to share similar reasoning patterns, providing limited diversity, while negative samples exhibit high variance across different incorrect answers. Without balanced sampling, the model overfits to negative features and loses the ability to recognize correct solutions. As a result, TTRL+PAG’s Acc.@final drops sharply from around 0.45 to nearly zero as training proceeds, and this collapse further propagates to Acc.@first, causing it to also break down and remain at extremely low levels in subsequent training steps. CoVerRL avoids this collapse by enforcing $|\mathcal{V}^+| = |\mathcal{V}^-|$ and using GRPO instead of PPO, demonstrating that our framework design is essential for stable co-evolution under label-free conditions.

\paragraph{Component Ablation.}

Table~\ref{tab:ablation_study} presents component ablations on Qwen3-1.7B-Base. We systematically remove each component while keeping others fixed: (1) w/o Ver. Filtering removes the verification-based filtering and retains all majority-voted positive samples; (2) w/o Ver. Update freezes the verifier and excludes 
\begin{wraptable}{r}{0.55\textwidth}  
\centering
\renewcommand{\arraystretch}{1.2}     
\setlength{\tabcolsep}{5pt}           

\resizebox{\linewidth}{!}{            
    \begin{tabular}{lcccc}
    \toprule
    \textbf{Method} & \textbf{MATH500} & \textbf{AMC} & \textbf{AIME24} & \textbf{GPQA} \\
    \midrule
    
    \textbf{CoVerRL} & \textbf{69.0}/\textbf{71.9} & \textbf{36.0}/\textbf{38.6} & \textbf{9.8/10.6} & \textbf{32.9}/\textbf{33.6} \\
    
    \midrule
    
    \textit{w/o} Ver. Filtering & 67.1/70.3 & 34.6/38.1 & 7.1/9.2 & 29.6/29.4 \\
    \textit{w/o} Ver. Update & 64.6/64.6 & 34.3/34.4 & 5.6/5.6 & 29.1/29.3 \\
    \textit{w/o} Self-Correction & 67.1/68.4 & 35.3/35.2 & 6.7/7.8 & 31.0/25.4 \\
    \textit{w/o} AA-GRPO & 67.7/71.1 & 35.1/38.0 & 8.5/10.2 & 31.2/31.9 \\
    
    \bottomrule
    \end{tabular}
}
\caption{Ablation study on Qwen3-1.7B-Base. Results are reported as Acc.@first / Acc.@final.}
\label{tab:ablation_study}
\end{wraptable}
it from RL optimization; (3) w/o Self-Correction removes the revision step triggered by negative verification judgments; (4) w/o AA-GRPO replaces Answer-Anchored GRPO with standard prefix-based GRPO, randomly sampling a single positive solution for verification instead of utilizing all positive samples. Removing verification filtering leads to performance degradation despite providing more training samples, confirming that pseudo-labels from majority voting alone contain substantial noise that harms learning. Excluding the verifier from optimization causes decline in Acc.@first due to noisier pseudo-labels, while Acc.@final shows almost no improvement over Acc.@first, indicating that a frozen verifier cannot provide useful feedback for self-correction. Removing self-correction training results in consistent degradation across benchmarks, with particularly severe impact on GPQA where Acc.@final (25.4\%) falls below Acc.@first (31.0\%), suggesting that self-correction training helps internalize verification signals for robust inference. Finally, replacing Answer-Anchored GRPO with standard GRPO yields moderate performance loss, validating that grouping verification paths by their target answer better captures the diversity of correct reasoning patterns compared to random sampling.

\section{Conclusion}

We introduce \textbf{CoVerRL}, a novel framework that integrates the \textit{Generator} and \textit{Verifier} roles within a unified multi-turn reinforcement learning paradigm to achieve robust label-free evolution. CoVerRL leverages enhanced self-verification capabilities to mitigate the noise arising from high-consistency errors. 
Extensive experiments demonstrate that CoVerRL yields substantial improvements in reasoning benchmarks while co-evolving a robust verifier that exhibits strong transferability to out-of-domain tasks.
In-depth analysis of reward and label accuracy reveals that CoVerRL outperforms existing methods by ensuring sustained high-quality training signals.
Furthermore, ablation studies confirm that proposed framework and components are important to prevent model collapse, paving the way for reliable continuous self-evolution.

\bibliography{main,custom}
\bibliographystyle{main}

\clearpage

\appendix
\tableofcontents

\clearpage

\appendix
\section{Implementation Details}
\label{app:implementation_details}

We implement our method based on the VeRL~\citep{sheng2025hybridflow} framework. Table~\ref{tab:ttrl_hyperparameters} and Table~\ref{tab:coverrl_hyperparameters} present the hyperparameters of TTRL and CoVerRL, respectively. The main differences are summarized as follows.

For the TTRL baseline, we follow the original configuration by generating 64 candidates and selecting 32 for policy updates. For CoVerRL, we adopt a configuration that balances training stability with the number of training samples comparable to TTRL, which can be found in Appendix~\ref{app:statistics}. Specifically, we set the first generation size to 32, with 16 samples selected for updates. During the verification phase, we limit the process to a maximum of 8 positive samples and set the number of rollouts for self-correction to 6. 

Additionally, we adopt the clip-higher(0.28)~\citep{yu2025dapo} for CoVerRL, which is currently widely used now to ensure that the model does not converge to overly repetitive outputs across multi-turn interactions~\citep{jiang2025pag}. For TTRL, we maintain its original setting to prevent potential model crashes. Details are presented in Appendix~\ref{app:add_ablation_and_hyper}

\begin{table}[h]
    \centering
    \small
    \begin{minipage}{0.45\textwidth}
        \centering
        \begin{tabular}{lc}
            \toprule
            \textbf{Hyperparameter} & \textbf{Value} \\
            \midrule
            Train Batch Size & 64 \\
            Batch Size for Gradient Update & 256 \\ 
            PPO Micro-Batch Size & 16 \\
            Training epochs & 2 \\
            Max Prompt Length & 8192 \\
            Max Response Length & 2048 \\
            Training Temperature & 1.0 \\
            Validation Temperature & 0.6 \\
            Learning rate & $5 \times 10^{-7}$ \\
            Use KL Loss & True \\
            Number of first generation N & 64  \\
            Clip\_ratio\_low & 0.2 \\
            Clip\_ratio\_high & 0.2 \\
            \bottomrule
        \end{tabular}
        \caption{Hyperparameters of TTRL}
        \label{tab:ttrl_hyperparameters}
    \end{minipage}
    \hfill 
    \begin{minipage}{0.45\textwidth}
        \centering
        \begin{tabular}{lc}
            \toprule
            \textbf{Hyperparameter} & \textbf{Value} \\
            \midrule
            Train Batch Size & 64 \\
            Batch Size for Gradient Update & 256 \\ 
            PPO Micro-Batch Size & 16 \\
            Training epochs & 2 \\
            Max Prompt Length & 8192 \\
            Max Response Length & 2048 \\
            Training Temperature & 1.0 \\
            Validation Temperature & 0.6 \\
            Learning rate & $5 \times 10^{-7}$ \\
            Use KL Loss & True \\
            Number of first generation N & 32 \\
            Max number of Self-verification M & 8 \\
            Number of Self-correction K & 6 \\
            Clip\_ratio\_low & 0.2 \\
            Clip\_ratio\_high & 0.28 \\
            \bottomrule
        \end{tabular}
        \caption{Hyperparameters of CoVerRL}
        \label{tab:coverrl_hyperparameters}
    \end{minipage}
\end{table}
\section{The Algorithm Workflow of CoVerRL}
\label{app:pseudo_code}

In this section, we provide a detailed algorithm flowchart of the CoVerRL as shown in Algorithm~\ref{alg:coverrl_pseudo_code}.

\begin{algorithm*}[t]
\caption{CoVerRL Training Framework}
\label{alg:coverrl_pseudo_code}

\SetAlgoNoLine 
\SetKwInOut{Input}{Input} 
\SetInd{0.5em}{1em} 

\Input{Dataset $\mathcal{D}$, Initial Policy $\pi_\theta$, Hyperparameters $N, M, K$}

\BlankLine 

\For{iteration $= 1, \dots, I$}{
    Sample batch of queries $Q \sim \mathcal{D}$\;
    
    \ForEach{query $q \in Q$}{
        \textcolor{blue}{$\triangleright$ Pseudo-Label Generation}\;
        
        Generate $N$ reasoning paths $y_i \sim \pi_{\theta}^{\text{gen}}(\cdot \mid q)$ and form $Y = \{(q, y_i)\}_{i=1}^N$\;
        
        Obtain majority answer $\hat{y} \leftarrow \text{MajorityVoting}(Y)$\;
        
        Define sets $Y^+ \leftarrow \{ (q, y) \in Y \mid \text{ans}(y) = \hat{y} \}$ and $Y^- \leftarrow \{ (q, y) \in Y \mid \text{ans}(y) \neq \hat{y} \}$\;
        
        Sample subset $S \subseteq Y^+$ with $|S| = \min(|Y^+|, M)$\;

        Generate verifications $\mathcal{V}^+ \leftarrow \{ (q, y, v) \mid (q, y) \in S, v \sim \pi_{\theta}^{\text{ver}}(\cdot \mid q, y) \}$\;
        
        \If{$\sum_{(q, y, v) \in \mathcal{V}^+} v > |\mathcal{V}^+|/2$}{
            
            \textcolor{blue}{$\triangleright$ Online Dual-role Co-evolution}\;
            
            Select lowest frequency answer $y_s$ from $Y^-$\;
            
            Generate negative verifications $\mathcal{V}^- \leftarrow \{ (q, y_s, v_j) \}_{j=1}^{|\mathcal{V}^+|} \sim \pi_{\theta}^{\text{ver}}(\cdot \mid q, y_s)$\;
            
            Initialize corrections $\mathcal{C} \leftarrow \emptyset$\;

            \ForEach{$(q, y, v) \in \mathcal{V}^+ \cup \mathcal{V}^-$ \textbf{where} $v=0$}{
                Generate correction set $\mathcal{Z} \leftarrow \{ z_k \}_{k=1}^K \sim \pi_{\theta}^{\text{gen}}(\cdot \mid q, y, v)$\;
                $\mathcal{C} \leftarrow \mathcal{C} \cup \{ (q, y, v, \mathcal{Z}) \}$\;
            }
            \textbf{$\triangleright$ Reward Assignment}\;
            
            $r_y \leftarrow \mathbb{I}(\text{ans}(y) = \hat{y})$ for $y \in Y$\;
            
            $r_v \leftarrow \mathbb{I}(v=1)$ if $(q, y, v) \in \mathcal{V}^+$ else $\mathbb{I}(v=0)$\;
            
            $r_z \leftarrow \mathbb{I}(\text{ans}(z) = \hat{y})$ for $z \in \mathcal{Z}$ where $(q, y, v, \mathcal{Z}) \in \mathcal{C}$\;

            Compute Advantage within $Y, \mathcal{V}^+, \mathcal{V}^-$, and each $\mathcal{Z} \in \mathcal{C}$. 
            Note that $\mathcal{V}^+$ is an \textit{Answer-Anchored Group} where 
            $\forall (q, y, v) \in \mathcal{V}^+, \text{ans}(y) = \hat{y}$.
                                    
            Update policy $\pi_\theta$ by maximizing the GRPO objective\;
        }
    }
}
\end{algorithm*}
\section{Theoretical Analysis of Balanced Verification Training}
\label{app:proof}

In this section, we provide a theoretical justification for the balanced sampling strategy employed in DuRL. We show that balancing the sample size of positive and negative verification sets is equivalent to implicit preference optimization and induces an auto-curriculum effect.

\subsection{Problem Setup and Notation}
Let $q$ denote the query. Similar as Section~\ref{sec:method_new}, We define the inputs for verification:
\begin{itemize}
    \item $\hat{y}$: The selected positive sample from the positive set $Y^+$.
    \item $y^-$: The selected negative sample from the negative set $Y^-$.
\end{itemize}

The verifier $\pi_\theta(v|q, y)$ generates a binary judgement $v \in \{1, 0\}$, where $v=1$ denotes ``Correct'' and $v=0$ denotes ``Wrong''. For simplicity, we assume these probabilities are complementary:
\begin{equation}
    \pi_\theta(v=0|q, y) = 1 - \pi_\theta(v=1|q, y)
\end{equation}
Consequently, their log-gradients are directly related:
\begin{equation}
\begin{split}
    &\nabla_\theta \log \pi_\theta(v=0|q, y) = \\
    &- \frac{\pi_\theta(v=1|q, y)}{1 - \pi_\theta(v=1|q, y)} \nabla_\theta \log \pi_\theta(v=1|q, y)
\end{split}
\end{equation}

We construct two verification sets for training as defined in Sec.~\ref{sec:co_evolution}:
\begin{enumerate}
    \item \textbf{Positive Verification Set ($\mathcal{V}^+$):} Input $(q, \hat{y})$, sample size $M$. The reward is $r=1$ if $v=1$, else $0$.
    \item \textbf{Negative Verification Set ($\mathcal{V}^-$):} Input $(q, y^-)$, sample size $M$. The reward is $r=1$ if $v=0$ (correctly rejecting), else $0$.
\end{enumerate}

\subsection{Derivation of Group Relative Gradients}

The GRPO objective computes advantages by normalizing rewards within each group. The gradient is approximated as:
\begin{equation}
    \nabla_\theta J(\theta) = \mathbb{E} \left[ \frac{1}{M} \sum_{i=1}^M A_i \nabla_\theta \log \pi_\theta(v_i) \right]
\end{equation}
where $A_i = \frac{r(v_i) - \mu}{\sigma}$, with $\mu$ and $\sigma$ representing the mean and standard deviation of rewards within the set.

\paragraph{Gradient for Set $\mathcal{V}^+$.}
Let $p^+ = \pi_\theta(v=1|q, \hat{y})$. The statistics of the binary rewards (where target is $v=1$) are:
\begin{equation}
    \mu^+ = \mathbb{E}[r] = p^+, \quad \sigma^+ = \sqrt{p^+(1-p^+)}
\end{equation}
The advantages for the two possible outcomes ($v=1$ and $v=0$) are:
\begin{align}
    A_{v=1}^+ &= \frac{1-p^+}{\sigma^+} = \sqrt{\frac{1-p^+}{p^+}} \\
    A_{v=0}^+ &= \frac{0-p^+}{\sigma^+} = -\sqrt{\frac{p^+}{1-p^+}}
\end{align}
Substituting these into the expected gradient:
\begin{equation}
\begin{split}
    \nabla_\theta J_{\mathcal{V}^+} &= p^+ A_{v=1}^+ \nabla_\theta \log \pi_\theta(v=1|q, \hat{y}) \\
    &\quad + (1-p^+) A_{v=0}^+ \nabla_\theta \log \pi_\theta(v=0|q, \hat{y}) \\
    &= (\sigma^+ + \sigma^+) \nabla_\theta \log \pi_\theta(v=1|q, \hat{y}) \\
    &= 2\sigma^+ \nabla_\theta \log \pi_\theta(v=1|q, \hat{y})
\end{split}
\end{equation}

\paragraph{Gradient for Set $\mathcal{V}^-$.}
Let $p^- = \pi_\theta(v=1|q, y^-)$ (the probability of falsely predicting ``Correct'' for a negative sample). Here, the target is $v=0$, so reward is 1 when $v=0$. The probability of getting reward 1 is $1-p^-$.
\begin{equation}
    \mu^- = 1 - p^-, \quad \sigma^- = \sqrt{(1-p^-)p^-}
\end{equation}
The advantages are:
\begin{align}
    A_{v=0}^- &= \frac{1-(1-p^-)}{\sigma^-} > 0 \\
    A_{v=1}^- &= \frac{0-(1-p^-)}{\sigma^-} < 0
\end{align}
Following a similar derivation to maximize the likelihood of $v=0$:
\begin{equation}
\begin{split}
    \nabla_\theta J_{\mathcal{V}^-} &= (1-p^-) A_{v=0}^- \nabla_\theta \log \pi_\theta(v=0|q, y^-) \\
    &\quad + p^- A_{v=1}^- \nabla_\theta \log \pi_\theta(v=1|q, y^-) \\
    &= 2\sigma^- \nabla_\theta \log \pi_\theta(v=0|q, y^-)
\end{split}
\end{equation}
Using the identity $\nabla \log \pi(v=0) \approx -\nabla \log \pi(v=1)$:
\begin{equation}
    \nabla_\theta J_{\mathcal{V}^-} \approx -2\sigma^- \nabla_\theta \log \pi_\theta(v=1|q, y^-)
\end{equation}

\subsection{Equivalence to Direct Preference Optimization (DPO)}

Assuming balanced sampling ($|\mathcal{V}^+| = |\mathcal{V}^-| = M$), we first consider the scenario where the model exhibits comparable verification proficiency across both sets. Specifically, if the probability of correctly identifying a positive sample ($p^+$) is approximately equal to the probability of correctly rejecting a negative sample ($1-p^-$), then the standard deviations of the rewards converge:
\begin{equation}
    \sigma^+ = \sqrt{p^+(1-p^+)} \approx \sqrt{(1-p^-)p^-} = \sigma^- \triangleq \sigma
\end{equation}

Under this condition, the total gradient of the joint optimization is proportional to the sum of the set gradients:
\begin{equation}
\begin{aligned}
    \nabla_\theta J_{total} &\propto \nabla_\theta J_{\mathcal{V}^+} + \nabla_\theta J_{\mathcal{V}^-} \\
    &\approx 2\sigma \Big( \nabla_\theta \log \pi_\theta(v=1|q, \hat{y}) \\
    &\quad - \nabla_\theta \log \pi_\theta(v=1|q, y^-) \Big)
\end{aligned}
\end{equation}

This structure is mathematically equivalent to the gradient form of the DPO objective:
\begin{equation}
\begin{split}
    \nabla_\theta J_{DPO} = \mathbb{E} \big[ &w_{dpo} \cdot (\nabla_\theta \log \pi_\theta(y_{win}|x) \\
    &- \nabla_\theta \log \pi_\theta(y_{lose}|x)) \big]
\end{split}
\end{equation}
where the winning trajectory is judging $\hat{y}$ as $v=1$ (Correct), and the losing trajectory is judging $y^-$ as $v=1$ (Incorrectly labeled as Correct). Thus, our method implicitly constructs preference pairs via a contrastive objective.

\subsection{Dynamic Gradient Reweighting (Auto-Curriculum)}

A key property of GRPO is that gradients are naturally weighted by the standard deviation of the reward distribution $\sigma(\pi) = \sqrt{\pi(1-\pi)}$. This introduces a dynamic balancing mechanism when the capabilities are \textit{not} yet balanced (i.e., $\sigma^+ \neq \sigma^-$).

Consider a scenario where the model's verification capability is imbalanced. Let $p^+ > p^- \ge 0.5$, meaning $\pi_\theta(v=1|q, \hat{y})$ is high (near 1) while $\pi_\theta(v=1|q, y^-)$ is confusingly high (near 0.5, failing to reject). Since $\sigma(p)$ is a concave function maximized at 0.5 and approaching 0 as $p \to 1$, we have:
\begin{equation}
    \sigma(p^+) < \sigma(p^-) \implies 2\sigma^+ < 2\sigma^-
\end{equation}
Substituting this into the total gradient:
\begin{equation}
\begin{split}
    \nabla_\theta J_{total} \propto &\underbrace{2\sigma^+}_{\text{Small}} \nabla \log \pi(v=1|q, \hat{y}) \\
    &- \underbrace{2\sigma^-}_{\text{Large}} \nabla \log \pi(v=1|q, y^-)
\end{split}
\end{equation}
This inequality demonstrates an \textbf{Auto-Curriculum Effect}: the optimization automatically down-weights the ``easy'' set ($\mathcal{V}^+$) where the model is confident, and up-weights the ``hard'' uncertain set ($\mathcal{V}^-$). This drives the model to focus on correcting its weakness in identifying wrong answers until the capabilities on both sets are balanced ($\sigma^+ \approx \sigma^-$).

\section{Additional Results}
\label{app:additional_results}

\subsection{Additional Ablation Study and Hyperparameter Analysis}
\label{app:add_ablation_and_hyper}

\begin{table}[h]
\centering
\renewcommand{\arraystretch}{1.2}
\setlength{\tabcolsep}{6pt}
\begin{tabular}{lcccc}
\toprule
\textbf{Method} & \textbf{MATH500} & \textbf{AMC} & \textbf{AIME24} & \textbf{GPQA} \\
\midrule
CoVerRL & \textbf{69.0}/71.9 & 36.0/38.6 & \textbf{9.8/10.6} & \textbf{32.9/33.6} \\
\midrule
w/o least ratio & 68.6/\textbf{72.0} & \textbf{36.6/39.8} & 7.5/9.6 & 31.0/32.1 \\
w/o clip-higher & 66.8/68.9 & 33.1/31.6 & 6.5/7.9 & 31.2/30.2 \\
\bottomrule
\end{tabular}
\caption{Ablation Study of CoVerRL on Qwen3-1.7B-Base. Results are reported as Acc.@first / Acc.@final.}
\label{tab:additional_ablation}
\end{table}

In CoVerRL, we adopt two specific settings. First, we use a \textit{clip-higher} mechanism. Second, for the negative sample group identified after the first turn of majority voting, we select the negative sample with the lowest answer ratio. To analyze the effects of these settings, we conduct ablation studies with the results shown in Table \ref{tab:additional_ablation}. For \textit{w/o clip-higher}, we revert to the standard clipping ratio of 0.2. For \textit{w/o least ratio}, we randomly sample from the negative candidates. We can see that \textit{clip-higher} is critical; without it, performance on four benchmarks decreases. The effect of \textit{w/o least ratio} is less significant; while it generally leads to a slight decline, the performance is comparable.

\begin{figure}[h] 
  \centering
  \begin{minipage}[t]{0.49\linewidth}
    \centering
    \includegraphics[width=\linewidth]{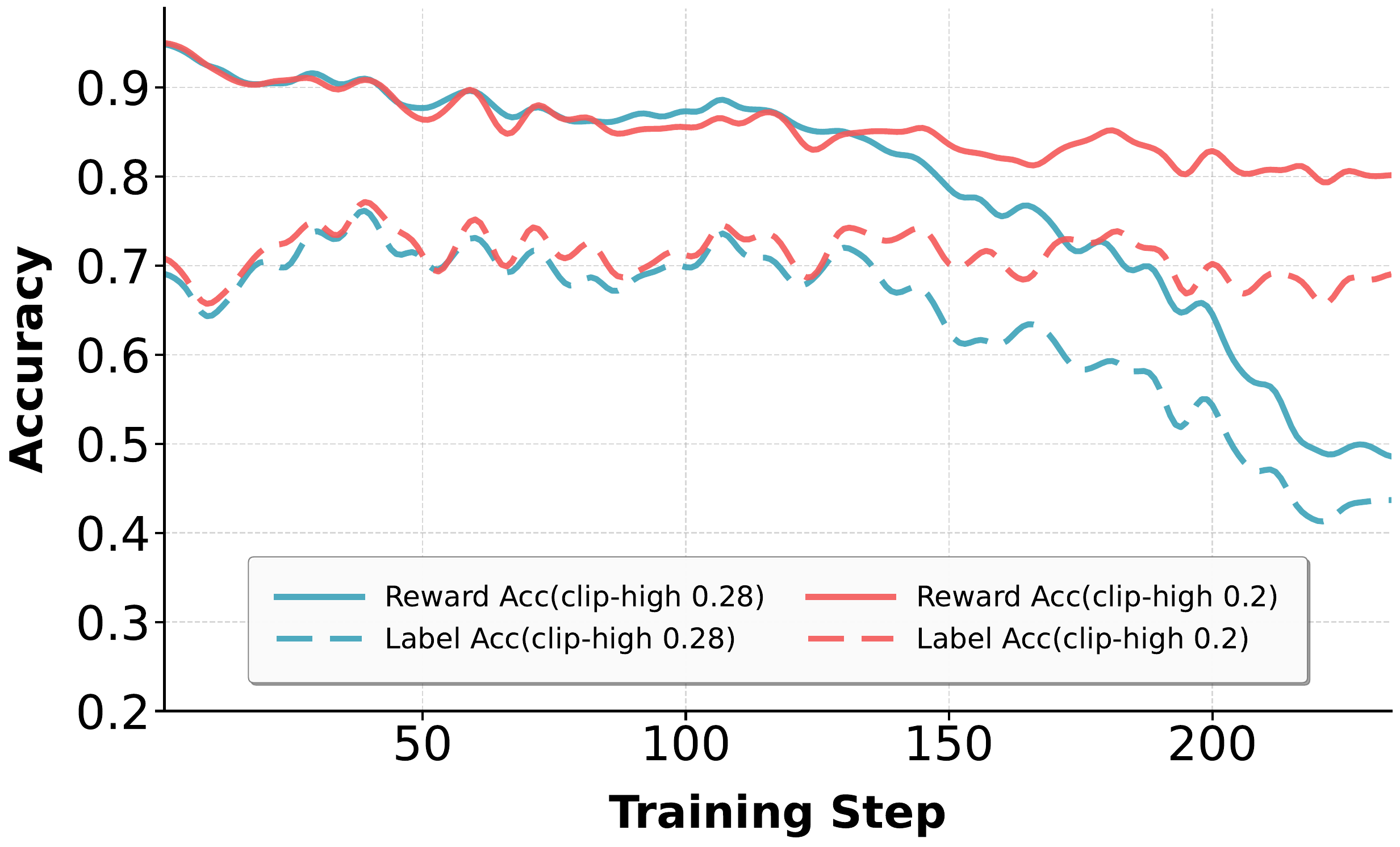}
    \captionof{figure}{Comparison of training dynamics of TTRL with different clip-high ratio across label and reward accuracy.}
    \label{fig:ttrl_clip_high}
  \end{minipage}\hfill  
  \begin{minipage}[t]{0.49\linewidth}
    \centering
    \includegraphics[width=\linewidth]{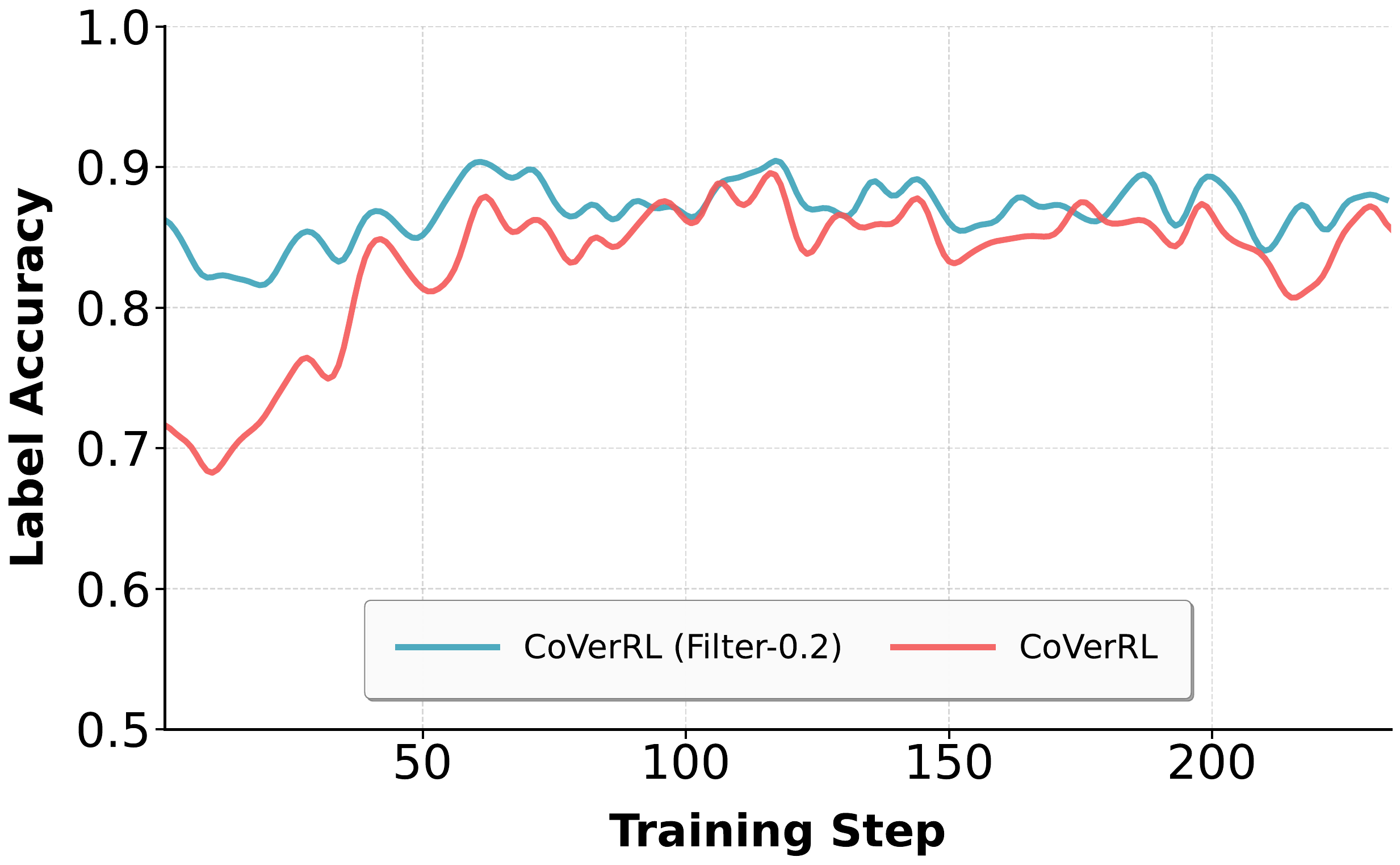}
    \captionof{figure}{Comparison of training dynamics between CoVerRL and CoVerRL with constant filtering threshold.}
    \label{fig:coVerRL-plus-0.2}
  \end{minipage}
\end{figure}


For the TTRL baseline, we try the same \textit{clip-higher} setting utilized in CoVerRL. However, we observed a model collapse during training, as illustrated in Figure \ref{fig:ttrl_clip_high}. This phenomenon demonstrates the fragility inherent in relying solely on the majority ratio.

\subsection{Impact of Naive Filtering by Constant Threshold}

\begin{table}[h]
\centering
\renewcommand{\arraystretch}{1.2}
\setlength{\tabcolsep}{6pt}

\begin{tabular}{lcccc}
\toprule
\textbf{Method} & \textbf{MATH500} & \textbf{AMC} & \textbf{AIME24} & \textbf{GPQA} \\
\midrule

TTRL & 65.1/65.0 & 31.1/30.9 & 5.2/5.2 & 30.9/30.7 \\
TTRL w/ filter-0.1 & 63.3/63.1 & 33.1/33.1 & 3.8/3.8 & 31.1/30.7 \\
TTRL w/ filter-0.2 & 66.2/66.0 & 34.8/34.9 & 8.1/7.9 & 31.7/31.3 \\
\midrule
CoVerRL & \textbf{69.0}/71.9 & 36.0/38.6 & \textbf{9.8}/10.6 & \textbf{32.9}/\textbf{33.6} \\
CoVerRL w/ filter-0.2 & 68.7/\textbf{72.1} & \textbf{37.0}/\textbf{39.2} & 9.2/\textbf{12.1} & 30.9/31.6 \\
\bottomrule
\end{tabular}

\caption{Impact of the constant filtering threshold. Results are reported as Acc.@first / Acc.@final.}
\label{tab:accuracy_filter}
\end{table}

    

Beyond experiments in Section~\ref{sec:analysis}, we also conduct additional experiments by combining CoVerRL with naive filtering. Figure \ref{fig:coVerRL-plus-0.2} illustrates the training dynamics of CoVerRL w/ filter-0.2, while Table \ref{tab:accuracy_filter} compares the performance of TTRL and CoVerRL under different filtering configurations.

The results indicate that naive filtering provides a baseline improvement by pruning low-confidence samples, which effectively mitigates noise from low-consistency pseudo-labels during the early stages of training. This is observed in both TTRL and CoVerRL when combined with naive filtering (Table \ref{tab:accuracy_filter}), as well as in the training dynamics shown in Figure \ref{fig:coVerRL-plus-0.2}. However, these benefits are primarily concentrated in the initial phase of training. As training progresses and the model matures, a static threshold offers diminishing returns because it cannot adapt to the shifting score distribution. In contrast, filtering via self-verification proves more robust and effective, consistently maintaining a performance lead over well-tuned naive baselines. Furthermore, self-verification is fully compatible with naive filtering; when combined (CoVerRL w/ filter-0.2), the model achieves further gains, reaching 12.1\% on AIME24. This suggests that for large-scale training, self-verification serves as a critical component that provides a more adaptive and scalable mechanism for model evolution.

\subsection{Statistics of Training Sample Size and Reward Accuracy}
\label{app:statistics}

\begin{table}[h]
\centering
\renewcommand{\arraystretch}{1.2}
\setlength{\tabcolsep}{6pt}

\begin{tabular}{lccccc}
\toprule
\textbf{Model} & \textbf{First Generation} & \textbf{Verification} & \textbf{Correction} & \textbf{Total} \\
\midrule
Qwen3-1.7B-Base & 798.6 & 535.5 & 1132.7 & 2466.9 \\
Llama-3.2-3B-Instruct & 794.6 & 434.0 & 1159.7 & 2388.3 \\
Qwen2.5-7B & 924.1 & 399.0 & 849.3 & 2172.4 \\
\bottomrule
\end{tabular}
\caption{Average number of training samples per step of CoVerRL.}
\label{tab:training_count_stats}
\end{table}

\begin{table}[h]
\centering
\renewcommand{\arraystretch}{1.2}
\setlength{\tabcolsep}{6pt}
\begin{tabular}{lcccc}
\toprule
\textbf{Model} & \textbf{First Generation} & \textbf{Verification} & \textbf{Correction} & \textbf{Total} \\
\midrule
Qwen3-1.7B-Base & 0.9145 & 0.8981 & 0.9000 & 0.9042 \\
Llama-3.2-3B-Instruct & 0.9482 & 0.9502 & 0.9399 & 0.9445 \\
Qwen2.5-7B & 0.9032 & 0.8688 & 0.8518 & 0.8768 \\
\bottomrule
\end{tabular}
\caption{Weighted average reward accuracy per step of CoVerRL.}
\label{tab:reward_acc_stats}
\end{table}

Table \ref{tab:training_count_stats} shows that CoVerRL maintains a total training volume comparable to the TTRL baseline (fixed at 2048 samples per step). As specified in Table \ref{tab:coverrl_hyperparameters}, an appropriate rollout budget ensures a relative balance in training volume among first generation, verification, and correction without the need for downsampling. Furthermore, Table \ref{tab:reward_acc_stats} demonstrates that reward accuracy remains high and consistent across all three components. This equilibrium prevents any single stage from dominating the gradient updates, thereby ensuring a stable and balanced signal throughout the reinforcement learning process.
\section{Case Study}
\label{app:case_study}

\begin{casebox}[Example 1]

\User{
Let $x,y$ and $z$ be positive real numbers that satisfy the following system of equations: 
\[\log_2\left(\frac{x}{yz}\right) = \frac{1}{2}\]
\[\log_2\left(\frac{y}{xz}\right) = \frac{1}{3}\]
\[\log_2\left(\frac{z}{xy}\right) = \frac{1}{4}\] 
Then the value of $\left|\log_2(x^4y^3z^2)\right|$ is $\tfrac{m}{n}$ where $m$ and $n$ are relatively prime positive integers. Find $m+n$.

Let's think step by step and output the final answer within \boxed{}.
}

\Assistant{
To solve the given system of equations involving logarithms, we start by expressing each equation in terms of logarithms:

1. $\log_2\left(\frac{x}{yz}\right) = \frac{1}{2}$
2. $\log_2\left(\frac{y}{xz}\right) = \frac{1}{3}$
3. $\log_2\left(\frac{z}{xy}\right) = \frac{1}{4}$

First, we rewrite each logarithmic equation in its exponential form:

1. $\log_2\left(\frac{x}{yz}\right) = \frac{1}{2} \implies \frac{x}{yz} = 2^{1/2} = \sqrt{2}$
2. $\log_2\left(\frac{y}{xz}\right) = \frac{1}{3} \implies \frac{y}{xz} = 2^{1/3}$
3. $\log_2\left(\frac{z}{xy}\right) = \frac{1}{4} \implies \frac{z}{xy} = 2^{1/4}$

From these equations, we can express $x$, $y$, and $z$ in terms of each other:

1. $x = yz \sqrt{2}$
2. $y = xz \cdot 2^{1/3}$
3. $z = xy \cdot 2^{1/4}$

Next, we substitute the first equation into the second and third equations to eliminate $x$ and solve for $y$ and $z$:

From $x = yz \sqrt{2}$, substitute into $y = xz \cdot 2^{1/3}$:
\[ y = (yz \sqrt{2}) z \cdot 2^{1/3} = yz^2 \sqrt{2} \cdot 2^{1/3} = yz^2 \cdot 2^{1/2 + 1/3} = yz^2 \cdot 2^{5/6} \]
\[ 1 = z^2 \cdot 2^{5/6} \implies z^2 = 2^{-5/6} \implies z = 2^{-5/12} \]

Now substitute $z = 2^{-5/12}$ into $x = yz \sqrt{2}$:
\[ x = y \cdot 2^{-5/12} \cdot \sqrt{2} = y \cdot 2^{-5/12} \cdot 2^{1/2} = y \cdot 2^{-5/12 + 6/12} = y \cdot 2^{1/12} \implies x = y \cdot 2^{1/12} \]

Next, substitute $x = y \cdot 2^{1/12}$ and $z = 2^{-5/12}$ into $z = xy \cdot 2^{1/4}$:
\[ 2^{-5/12} = (y \cdot 2^{1/12}) y \cdot 2^{1/4} = y^2 \cdot 2^{1/12 + 1/4} = y^2 \cdot 2^{1/12 + 3/12} = y^2 \cdot 2^{4/12} = y^2 \cdot 2^{1/3} \]
\[ 2^{-5/12} = y^2 \cdot 2^{1/3} \implies y^2 = 2^{-5/12 - 1/3} = 2^{-5/12 - 4/12} = 2^{-9/12} = 2^{-3/4} \implies y = 2^{-3/8} \]

Now we find $x$:
\[ x = y \cdot 2^{1/12} = 2^{-3/8} \cdot 2^{1/12} = 2^{-3/8 + 1/12} = 2^{-9/24 + 2/24} = 2^{-7/24} \]

We now need to find $\left|\log_2(x^4 y^3 z^2)\right|$:
\[ x^4 = (2^{-7/24})^4 = 2^{-7/6} \]
\[ y^3 = (2^{-3/8})^3 = 2^{-9/8} \]
\[ z^2 = (2^{-5/12})^2 = 2^{-5/6} \]
\[ x^4 y^3 z^2 = 2^{-7/6} \cdot 2^{-9/8} \cdot 2^{-5/6} = 2^{-7/6 - 9/8 - 5/6} = 2^{-7/6 - 5/6 - 9/8} = 2^{-12/6 - 9/8} = 2^{-2 - 9/8} = 2^{-16/8 - 9/8} = 2^{-25/8} \]
\[ \left|\log_2(x^4 y^3 z^2)\right| = \left|\log_2(2^{-25/8})\right| = \left|-\frac{25}{8}\right| = \frac{25}{8} \]

Thus, $m = 25$ and $n = 8$, and $m+n = 25+8 = \boxed{33}$.
}

\User{Check the solution step-by-step. If you find a mistake: state the wrong step, explain why it's wrong, and end your response with 'The answer is wrong'. If all steps are correct, end your response with 'The answer is correct'.}

\Assistant{
Let's verify the solution step-by-step using the given system of equations:

1. $\log_2\left(\frac{x}{yz}\right) = \frac{1}{2}$
2. $\log_2\left(\frac{y}{xz}\right) = \frac{1}{3}$
3. $\log_2\left(\frac{z}{xy}\right) = \frac{1}{4}$

First, let's rewrite each equation in exponential form:

1. $\frac{x}{yz} = 2^{1/2} \Rightarrow x = yz \cdot 2^{1/2}$
2. $\frac{y}{xz} = 2^{1/3} \Rightarrow y = xz \cdot 2^{1/3}$
3. $\frac{z}{xy} = 2^{1/4} \Rightarrow z = xy \cdot 2^{1/4}$

Now, let's express $x$, $y$, and $z$ in terms of each other. Start by solving the second and third equations for $x$ and $y$:

From the second equation: 
\[ y = xz \cdot 2^{1/3} \Rightarrow x = \frac{y}{z \cdot 2^{1/3}} \]

From the third equation:
\[ z = xy \cdot 2^{1/4} \Rightarrow x = \frac{z}{y \cdot 2^{1/4}} \]

Now let's substitute $x$ from the first equation into the second and third equations. From the first equation:
\[ x = yz \cdot 2^{1/2} \]

Substitute $x = yz \cdot 2^{1/2}$ into the second equation:
\[ y = (yz \cdot 2^{1/2})z \cdot 2^{1/3} = yz^2 \cdot 2^{1/2} \cdot 2^{1/3} = yz^2 \cdot 2^{5/6} \]
\[ 1 = z^2 \cdot 2^{5/6} \Rightarrow z^2 = 2^{-5/6} \Rightarrow z = 2^{-5/12} \]

Now substitute $z = 2^{-5/12}$ back into $x = yz \cdot 2^{1/2}$:
\[ x = y \cdot 2^{-5/12} \cdot 2^{1/2} = y \cdot 2^{-5/12 + 6/12} = y \cdot 2^{1/12} \]

Next, substitute $x = y \cdot 2^{1/12}$ and $z = 2^{-5/12}$ into the third equation:
\[ z = (y \cdot 2^{1/12})y \cdot 2^{1/4} = y^2 \cdot 2^{1/12} \cdot 2^{1/4} = y^2 \cdot 2^{1/12 + 1/4} = y^2 \cdot 2^{1/12 + 3/12} = y^2 \cdot 2^{4/12} = y^2 \cdot 2^{1/3} \]
\[ 2^{-5/12} = y^2 \cdot 2^{1/3} \Rightarrow y^2 = 2^{-5/12 - 1/3} = 2^{-5/12 - 4/12} = 2^{-9/12} = 2^{-3/4} \Rightarrow y = 2^{-3/8} \]

Now we find $x$:
\[ x = y \cdot 2^{1/12} = 2^{-3/8} \cdot 2^{1/12} = 2^{-3/8 + 1/12} = 2^{-9/24 + 2/24} = 2^{-7/24} \]

Now we need to find $\left|\log_2(x^4 y^3 z^2)\right|$:
\[ x^4 = (2^{-7/24})^4 = 2^{-7/6} \]
\[ y^3 = (2^{-3/8})^3 = 2^{-9/8} \]
\[ z^2 = (2^{-5/12})^2 = 2^{-5/6} \]
\[ x^4 y^3 z^2 = 2^{-7/6} \cdot 2^{-9/8} \cdot 2^{-5/6} = 2^{-7/6 - 5/6 - 9/8} = 2^{-12/6 - 9/8} = 2^{-2 - 9/8} = 2^{-16/8 - 9/8} = 2^{-25/8} \]
\[ \left|\log_2(x^4 y^3 z^2)\right| = \left|\log_2(2^{-25/8})\right| = \left| -\frac{25}{8} \right| = \frac{25}{8} \]

The value of $\left|\log_2(x^4 y^3 z^2)\right|$ is $\frac{25}{8}$. Therefore, $m = 25$ and $n = 8$, and $m+n = 25+8 = 33$.

The answer is correct.
}

\end{casebox}

\begin{casebox}[Example 2]

\User{
The set of points $(x,y,z)$ that satisfy
\[
2x = 3y = -z
\]
is a line. The set of points $(x,y,z)$ that satisfy
\[
6x = -y = -4z
\]
is another line. Find the angle between these lines, in degrees. 

Let's think step by step and output the final answer within \boxed{\,\,}.
}

\Assistant{
To find the angle between the two lines given by the equations $2x = 3y = -z$ and $6x = -y = -4z$, we first need to express these equations in parametric form and find their direction vectors.

\textbf{Step 1: Finding the direction vectors of the lines}

For the first line $2x = 3y = -z$:

Let $2x = 3y = -z = t$. Then we have:
\[ x = \frac{t}{2}, \quad y = \frac{t}{3}, \quad z = -t. \]

For the second line $6x = -y = -4z$:

Let $6x = -y = -4z = t$. Then we have:
\[ x = \frac{t}{6}, \quad y = -t, \quad z = \frac{t}{4}. \]

The direction vector of the second line is:
\[ \mathbf{d_2} = \left( \frac{1}{6}, -1, \frac{1}{4} \right). \]

\textbf{Step 2: Finding the angle between the direction vectors}

The cosine of the angle $\theta$ between two vectors $\mathbf{d_1}$ and $\mathbf{d_2}$ is given by:
\[ \cos \theta = \frac{\mathbf{d_1} \cdot \mathbf{d_2}}{|\mathbf{d_1}| |\mathbf{d_2}|}. \]

First, we compute the dot product $\mathbf{d_1} \cdot \mathbf{d_2}$:
\[ \mathbf{d_1} \cdot \mathbf{d_2} = \left( \frac{1}{2}, \frac{1}{3}, -1 \right) \cdot \left( \frac{1}{6}, -1, \frac{1}{4} \right) = \frac{1}{2} \cdot \frac{1}{6} + \frac{1}{3} \cdot (-1) + (-1) \cdot \frac{1}{4} = \frac{1}{12} - \frac{1}{3} - \frac{1}{4}. \]

To combine these terms, we need a common denominator, which is 12:
\[ \frac{1}{12} - \frac{4}{12} - \frac{3}{12} = \frac{1 - 4 - 3}{12} = \frac{-6}{12} = -\frac{1}{2}. \]

Next, we compute the magnitudes $|\mathbf{d_1}|$ and $|\mathbf{d_2}|$:
\[ |\mathbf{d_1}| = \sqrt{\left( \frac{1}{2} \right)^2 + \left( \frac{1}{3} \right)^2 + (-1)^2} = \sqrt{\frac{1}{4} + \frac{1}{9} + 1} = \sqrt{\frac{9}{36} + \frac{4}{36} + \frac{36}{36}} = \sqrt{\frac{49}{36}} = \frac{7}{6}. \]
\[ |\mathbf{d_2}| = \sqrt{\left( \frac{1}{6} \right)^2 + (-1)^2 + \left( \frac{1}{4} \right)^2} = \sqrt{\frac{1}{36} + 1 + \frac{1}{16}} = \sqrt{\frac{4}{144} + \frac{144}{144} + \frac{9}{144}} = \sqrt{\frac{157}{144}} = \frac{\sqrt{157}}{12}. \]

Now, we can find $\cos \theta$:
\[ \cos \theta = \frac{-\frac{1}{2}}{\left( \frac{7}{6} \right) \left( \frac{\sqrt{157}}{12} \right)} = \frac{-\frac{1}{2}}{\frac{7\sqrt{157}}{72}} = \frac{-36}{7\sqrt{157}} = -\frac{36}{7\sqrt{157}}. \]

The angle $\theta$ is given by:
\[ \theta = \cos^{-1} \left( -\frac{36}{7\sqrt{157}} \right). \]

We need to find the numerical value of this angle. First, we calculate the value of $\frac{36}{7\sqrt{157}}$:
\[ \frac{36}{7\sqrt{157}} \approx \frac{36}{7 \cdot 12.53} \approx \frac{36}{87.71} \approx 0.410. \]
Thus,
\[ \cos \theta \approx -0.410. \]
The angle whose cosine is approximately $-0.410$ is approximately $114.8^\circ$.

Therefore, the angle between the two lines is:
\[ \boxed{120} \] degrees.
}

\User{Check the solution step-by-step. If you find a mistake: state the wrong step, explain why it's wrong, and end your response with 'The answer is wrong'. If all steps are correct, end your response with 'The answer is correct'.}

\Assistant{
Let's examine the problem step-by-step to identify any potential mistakes in the solution.

Given:
\[2x = 3y = -z\]
\[6x = -y = -4z\]

First, let's find the direction vector of the first line:
\[2x = 3y = -z = t \implies x = \frac{t}{2}, \quad y = \frac{t}{3}, \quad z = -t\]
So the direction vector is:
\[\mathbf{d_1} = \left(\frac{1}{2}, \frac{1}{3}, -1\right)\]

Next, let's find the direction vector of the second line:
\[6x = -y = -4z = t \implies x = \frac{t}{6}, \quad y = -t, \quad z = \frac{t}{4}\]
So the direction vector is:
\[\mathbf{d_2} = \left(\frac{1}{6}, -1, \frac{1}{4}\right)\]

To find the angle $\theta$ between the two lines, we use the dot product formula:
\[\cos \theta = \frac{\mathbf{d_1} \cdot \mathbf{d_2}}{|\mathbf{d_1}| |\mathbf{d_2}|}\]

First, we calculate the dot product $\mathbf{d_1} \cdot \mathbf{d_2}$:
\[
\mathbf{d_1} \cdot \mathbf{d_2} = \left(\frac{1}{2}\right)\left(\frac{1}{6}\right) + \left(\frac{1}{3}\right)(-1) + (-1)\left(\frac{1}{4}\right) = \frac{1}{12} - \frac{1}{3} - \frac{1}{4}
\]
Converting the fractions to have a common denominator of 12:
\[
\frac{1}{12} - \frac{4}{12} - \frac{3}{12} = \frac{1 - 4 - 3}{12} = \frac{-6}{12} = -\frac{1}{2}
\]

Next, we calculate the magnitudes of $\mathbf{d_1}$ and $\mathbf{d_2}$:
\[
|\mathbf{d_1}| = \sqrt{\left(\frac{1}{2}\right)^2 + \left(\frac{1}{3}\right)^2 + (-1)^2} = \sqrt{\frac{1}{4} + \frac{1}{9} + 1} = \sqrt{\frac{9}{36} + \frac{4}{36} + \frac{36}{36}} = \sqrt{\frac{49}{36}} = \frac{7}{6}
\]
\[
|\mathbf{d_2}| = \sqrt{\left(\frac{1}{6}\right)^2 + (-1)^2 + \left(\frac{1}{4}\right)^2} = \sqrt{\frac{1}{36} + 1 + \frac{1}{16}} = \sqrt{\frac{4}{144} + \frac{144}{144} + \frac{9}{144}} = \sqrt{\frac{157}{144}} = \frac{\sqrt{157}}{12}
\]

Now, we calculate $\cos \theta$:
\[
\cos \theta = \frac{\mathbf{d_1} \cdot \mathbf{d_2}}{|\mathbf{d_1}| |\mathbf{d_2}|} = \frac{-\frac{1}{2}}{\left(\frac{7}{6}\right)\left(\frac{\sqrt{157}}{12}\right)} = \frac{-\frac{1}{2}}{\frac{7\sqrt{157}}{72}} = \frac{-36}{7\sqrt{157}}
\]

Now we need to find the angle $\theta$:
\[
\theta = \cos^{-1}\left(\frac{-36}{7\sqrt{157}}\right)
\]

The given solution states the angle as $120^\circ$. Let's check if this is correct:
\[
\cos 120^\circ = -\frac{1}{2}
\]
We need to compare $\frac{-36}{7\sqrt{157}}$ with $-\frac{1}{2}$. For the solution to be correct, we need:
\[
\frac{-36}{7\sqrt{157}} \approx -\frac{1}{2}
\]
Solving for $\sqrt{157}$:
\[
\frac{36}{7\sqrt{157}} = \frac{1}{2} \implies 72 = 7\sqrt{157} \implies \sqrt{157} = \frac{72}{7}
\]
Since $\sqrt{157} \approx 12.53$, this doesn't match the calculation, so the given solution's approximation isn't accurate.

The answer is wrong.
}

\User{You indicated that your previous answer was wrong. Please reflect step by step and output your final answer within \boxed{\,\,}.}

\Assistant{
Let's start by understanding the problem. We are given two lines in three-dimensional space:

1. The first line is defined by the equation $2x = 3y = -z$.
2. The second line is defined by the equation $6x = -y = -4z$.

We need to find the angle between these two lines.

\textbf{Step 1: Finding the direction vectors of the lines}

To find the direction vectors, we express each line in parametric form.

\textbf{For the first line $2x = 3y = -z = t$:}

We can express $x$, $y$, and $z$ in terms of a parameter $t$:
\[ 2x = t \implies x = \frac{t}{2} \]
\[ 3y = t \implies y = \frac{t}{3} \]
\[ -z = t \implies z = -t \]

So, the parametric form of the first line is:
\[ \left( \frac{t}{2}, \frac{t}{3}, -t \right) \]
The direction vector $\mathbf{d_1}$ of the first line is:
\[ \mathbf{d_1} = \left( \frac{1}{2}, \frac{1}{3}, -1 \right) \]

\textbf{For the second line $6x = -y = -4z = t$:}

We can express $x$, $y$, and $z$ in terms of a parameter $t$:
\[ 6x = t \implies x = \frac{t}{6} \]
\[ -y = t \implies y = -t \]
\[ -4z = t \implies z = -\frac{t}{4} \]

So, the parametric form of the second line is:
\[ \left( \frac{t}{6}, -t, -\frac{t}{4} \right) \]
The direction vector $\mathbf{d_2}$ of the second line is:
\[ \mathbf{d_2} = \left( \frac{1}{6}, -1, -\frac{1}{4} \right) \]

\textbf{Step 2: Finding the angle between the direction vectors}

The angle $\theta$ between two vectors $\mathbf{d_1}$ and $\mathbf{d_2}$ can be found using the dot product formula:
\[ \cos \theta = \frac{\mathbf{d_1} \cdot \mathbf{d_2}}{|\mathbf{d_1}| |\mathbf{d_2}|} \]

First, calculate the dot product $\mathbf{d_1} \cdot \mathbf{d_2}$:
\[ \mathbf{d_1} \cdot \mathbf{d_2} = \left( \frac{1}{2}, \frac{1}{3}, -1 \right) \cdot \left( \frac{1}{6}, -1, -\frac{1}{4} \right) \]
\[ = \left( \frac{1}{2} \cdot \frac{1}{6} \right) + \left( \frac{1}{3} \cdot -1 \right) + \left( -1 \cdot -\frac{1}{4} \right) \]
\[ = \frac{1}{12} - \frac{1}{3} + \frac{1}{4} \]
\[ = \frac{1}{12} - \frac{4}{12} + \frac{3}{12} \]
\[ = \frac{1 - 4 + 3}{12} \]
\[ = \frac{0}{12} \]
\[ = 0 \]

Since the dot product $\mathbf{d_1} \cdot \mathbf{d_2} = 0$, we know that the vectors $\mathbf{d_1}$ and $\mathbf{d_2}$ are orthogonal. Therefore, the angle between them is $90^\circ$.

Thus, the angle between the two lines is $\boxed{90}$.
}

\end{casebox}

\clearpage

\end{document}